\documentclass[5p,times]{elsarticle}

\usepackage{amssymb}
\usepackage{amsmath}
\usepackage{hyperref}
\usepackage{algorithm}
\usepackage{algorithmicx}
\usepackage{algpseudocode}
\usepackage{booktabs}
\usepackage{float}
\usepackage{multirow}
\usepackage{microtype}
\usepackage{orcidlink}
\journal{Future Generation Computer Systems}

\begin{document}

\begin{frontmatter}



\title{From XAI to MLOps: Explainable Concept Drift Detection with Profile Drift Detection}


\author[label1]{Ugur Dar} %% Author name
\author[label1]{Mustafa Cavus \orcidlink{0000-0002-6134-0471}} %% Author name

\affiliation[label1]{organization={Eskisehir Technical University},
            addressline={Department of Statistics}, 
            city={Eskisehir},
            postcode={26555},
            country={Türkiye}}

%% Abstract
\begin{abstract}
Predictive models often degrade in performance due to evolving data distributions, a phenomenon known as data drift. Among its forms, concept drift—where the relationship between explanatory variables and the response variable changes—is particularly challenging to detect and adapt to. Traditional drift detection methods often rely on metrics such as accuracy or marginal variable distributions, which may fail to capture subtle but important conceptual changes. This paper proposes a novel method, \emph{Profile Drift Detection} (PDD), which enables both the detection of concept drift and an enhanced understanding of its underlying causes by leveraging an explainable AI tool—Partial Dependence Profiles (PDPs). PDD quantifies changes in PDPs through new drift metrics that are sensitive to shifts in the data stream while remaining computationally efficient. This approach is aligned with MLOps practices, emphasizing continuous model monitoring and adaptive retraining in dynamic environments. Experiments on synthetic and real-world datasets demonstrate that PDD outperforms existing methods by maintaining high predictive performance while effectively balancing sensitivity and stability in drift signals. The results highlight its suitability for real-time applications, and the paper concludes by discussing the method's advantages, limitations, and potential extensions to broader use cases.
\end{abstract}



%% Keywords

\end{frontmatter}


\section{Introduction}
\label{sec1}
\begin{figure}[h]
    \centering
    \includegraphics[width = \linewidth]{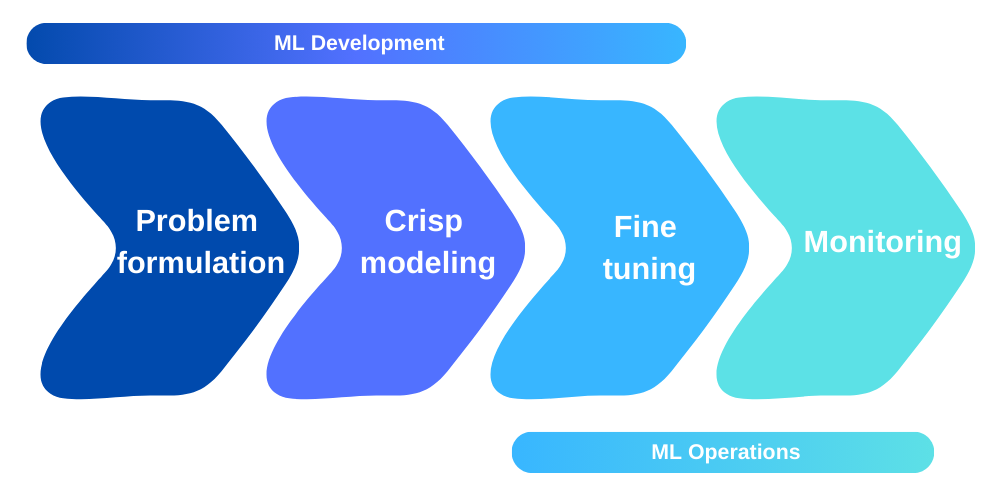}
    \caption{The predictive modeling process}
    \label{fig:mdp}
\end{figure}

The stationary and unbiased distribution assumption in predictive models states that observations in the training and test sets are independent and identically distributed \citep{cieslak_and_chawla_2009}. However, this assumption often does not hold in practice, leading to significant decreases in the performance of predictive models \citep{alaiz_et_al_2008}. Data characteristics can change over time, rendering a trained model obsolete and forcing it to adapt to these changes. The constantly evolving nature of data streams presents new challenges for predictive models \citep{moreno_et_al_2012}. These models must demonstrate high predictive accuracy and quickly incorporate new information while remaining computationally light and responsible \citep{korycki_and_krawczyk_2023}. Model monitoring, which is a part of MLOps, has become a necessary additional layer in the predictive modeling process in Figure~\ref{fig:mdp} to ensure that the ML model maintains consistent behavior over time in real-world applications \citep{biecek_2019,mougan_and_nielsen_2023}.

One of the problems that causes inconsistency in behavior is called data drift, whose types are also classified into three categories: (1) Covariate drift, (2) Label drift, and (3) Concept drift. Data drift may manifest in various forms, such as increased electricity consumption due to significant life changes like pandemic-related lockdowns \citep{fumagalli_et_al_2023}, changes in airline customer behaviors \citep{garg_et_al_2021}, and shifts in supply chain planning due to evolving consumer behavior \citep{kulinski_and_inouye_2023}. It is essential to detect these drifts on time to take precautions for the model.

Various methods are used to detect data drifts, categorized into three groups: statistical, sequential analysis-based, and win\-dow-ba\-sed methods. For detecting concept drift, which is characterized as a change in the relationship between explanatory variables and the response variable, there are instances where current methods in the literature fall short \citep{demvsar_and_bosnic_2018,duckworth_et_al_2021,rahmani_et_al_2023}. This limitation highlights the need for MLOps, which integrates model monitoring and adaptation strategies to address data drift effectively. MLOps practices focus on maintaining model performance and managing change, enabling predictive models to adapt to dynamic data environments in real time. One particularly challenging scenario is when concept drift occurs without performance and distribution metrics changes, rendering traditional detection methods unreliable. In response to this situation, some alternative approaches have been proposed where the methods used in the literature fail to detect concept drift adequately. 

Moreover, \cite{demvsar_and_bosnic_2018} emphasized that current data drift detection methods rely solely on performance and distribution metrics, which are insufficient to understand the underlying cause of changes in the model. They developed an alternative method using the local interpretable model-agnostic explanations \citep{strumbelj_et_al_2009} and demonstrated through comparative studies that it offers better detection performance than existing methods. \cite{duckworth_et_al_2021} used the normalized SHAP method \citep{lundberg_2017} to detect data drifts in emergency department length-of-stay prediction models following COVID-19, noting that traditional methods failed to reflect changes in the model's information utilization. \cite{muschalik_et_al_2022} developed an approach to detect concept drifts using permutation-based variable importance \citep{breiman2001random} values. However, they highlighted the need to address complexity, effectiveness, sensitivity, and explainability for their approach to be effectively applied. \cite{rahmani_et_al_2023} explored data drifts observed in models using the SHAP method for calculating the variable importance. Their most significant challenge was the inability to distinguish between covariate and concept drift, primarily due to the inappropriate use of XAI tools. While SHAP provides sufficient information on variable importance, it fails to detect the relationship between explanatory and response variables. \cite{mougan_and_nielsen_2023} emphasized that statistical tests operate solely based on the data distribution and may lead to false alarms. Their study developed an explainable concept drift detection method for regression problems using non-parametric bootstrap uncertainty estimation and SHAP values. \cite{mattos_et_al_2021} highlighted a lack of studies focusing on the causes and explainability of concept drift detection methods. They developed a technique based on a decision tree model to detect and visualize concept drift by identifying changes in the regions partitioned by the decision tree. However, they discussed the limitations of their approach, such as the cost of continuously training the model and the feasibility of tracking the limited number of variables selected by the model. Additionally, when non-tree-based models are used, the method might evaluate different relationships from those in the decision process, making it more suitable as a tree-based method. \cite{muschalik_et_al_2022} developed an approach for detecting concept drift using permutation-based variable importance values. Similarly, \cite{fumagalli_et_al_2023} developed an incremental permutation-based variable importance method for use in incremental models. However, they emphasized that for their approach to be effectively applied, it is necessary to address shortcomings in complexity, efficiency, sensitivity, and explainability.

There is a limited body of literature on data drift detection methods that employ XAI tools. Even when these methods are used, they often face challenges in some data drifts, or they rely on high computational cost methods such as permutation-based variable importance or SHAP. These approaches are not always practical for detecting changes in the relationship between the response variable and explanatory variables. Therefore, this paper introduces a solution based on the partial dependence profile (PDP), which is an XAI tool used to uncover the relationship between the response variable and explanatory variables. Because it is a versatile tool that is not only used to examine the relationship between the response variable and explanatory variables but also used in hyper-parameter optimization \citep{moosbauer_et_al_2021} and for determining variable importance and identifying interactions between variables \citep{inglis_et_al_2022}.

In addition to the scientific literature, software, and technologies that include data drift detection methods, which are complex to implement, they are quite limited. Although there are basic R tools such as the \href{https://cran.r-project.org/web/packages/vetiver/index.html}{vetiver} framework and \href{https://cran.r-project.org/package=pins}{pins} for monitoring model metrics, they do not provide specific instruments for data drift monitoring. On the other hand, \href{https://cran.r-project.org/package=drifter}{drifter} offers useful functions for detecting data drift based on distances between distributions, residuals, and PDPs of two models, but it is limited to comparing two trained models and is not suitable for streaming data. Meanwhile, \href{https://cran.r-project.org/package=harbinger}{harbinger} supplies some functions to detect anomalies, drifts, and change points, but is focused on time series data. While R and CRAN are immature in terms of model monitoring, there are many technologies and Python libraries available such as EvidentlyAI\footnote{\href{https://www.evidentlyai.com/}{https://www.evidentlyai.com/}}, SeldonIO\footnote{\href{https://www.seldon.io/}{https://www.seldon.io/}}, NannyML\footnote{\href{https://nannyml.readthedocs.io/en/stable/index.html}{https://nannyml.readthedocs.io/en/stable/index.html}}, Frouros\footnote{\href{https://frouros.readthedocs.io/en/latest/}{https://frouros.readthedocs.io/en/latest/}}, River\footnote{\href{https://riverml.xyz/0.8.0/examples/concept-drift-detection/}{https://riverml.xyz/0.8.0/examples/concept-drift-detection/}}.

To address these gaps, we propose a novel concept drift detection method, Profile Drift Detection (PDD), based on explainable AI techniques. PDD leverages Partial Dependence Profiles to monitor concept drifts in production environments. The key contributions of our study are threefold: First, we introduce PDD as an explainable drift detection method that quantifies changes in the relationship between variables and the target variable through three complementary metrics (PDI, L2, and L2Der). Second, through comprehensive experiments on synthetic and real-world datasets, we demonstrate that PDD effectively balances sensitivity and stability in drift detection while maintaining high predictive performance. Third, unlike black-box methods, PDD provides interpretable insights into which variables drive the drift and how their relationships with the target variable evolve.\footnote{An R implementation of PDD and baseline methods is available in the datadriftR package (\href{https://cran.r-project.org/package=datadriftR}{CRAN}). Source code and reproducibility materials: \url{https://github.com/ugurdar/datadriftR_DMKD}}
%%%%%%%%%%%%%%%%%%%%%%%%%%%%%%%%%%%%%%%%%%%%%%%%%%%%%%%%%%%%%%%%%%%%%%%%%%%%%

\section{Preliminaries}

Let $f_{\theta}: \mathcal{X} \rightarrow \mathcal{Y}$ be a predictive model, where $f_{\theta}$ represents the function learned by the model, $\theta$ denotes the parameters to be optimized, $\mathcal{X}$ is the input space, and $\mathcal{Y}$ is the output space. The model is trained on a dataset $D = \{(x_i, y_i)\}_{i=1}^{n}$, where $x_i \in \mathcal{X}$ are the explanatory variables and $y \in \mathcal{Y}$ is the response variable. The learning process aims to find the optimal parameters $\hat{\theta}$ by minimizing a loss function $L(f_{\theta}(x_i), y_i)$, which quantifies the difference between the model's predictions $f_{\theta}(x_i)$ and the true values $y_i$. The objective is to solve $\hat{\theta} = \arg \min_{\theta} \sum_{i=1}^{n} L(f_{\theta}(x_i), y_i)$, thereby achieving the best possible model performance on the given data. Data drift occurs when there is a discrepancy between the distribution of the data used to train a model, denoted as $P_{\text{train}}(X, Y)$, and the distribution of the data the model encounters during deployment or testing denoted as $P_{\text{test}}(X, Y)$. This drift can negatively impact the model's performance, as the model learns patterns based on the train data distribution.\\

\noindent \textbf{Covariate drift.} It occurs when the distribution of explanatory variables changes while the distribution of the response variable remains unchanged: 

\begin{equation}
P_{\text{train}}(X) \neq P_{\text{test}}(X).   
\end{equation}

\noindent In this case, the model might struggle to make accurate predictions on new data, even though it performed well on the train set.\\

\noindent \textbf{Label drift.} It indicates a change in the distribution of the response variable while the distribution of the explanatory variable stays constant: 

\begin{equation}
P_{\text{train}}(Y) \neq P_{\text{test}}(Y). 
\end{equation}

\noindent Label shift is particularly problematic in scenarios with imbalanced classes, as it can increase misclassification rates.\\

\noindent \textbf{Concept drift.} It occurs when the conditional relationship between explanatory variables and the response variable changes:

\begin{equation}
P_{\text{train}}(Y|X) \neq P_{\text{test}}(Y|X),
\end{equation}

\noindent even if the marginal distributions $P(X)$ and $P(Y)$ remain unchanged. This fundamentally alters how the model interprets input variables, leading to reduced accuracy and reliability.

\noindent There are several methods proposed to detect these kinds of drifts in the literature, which are described in the following section.
%%%%%%%%%%%%%%%%%%%%%%%%%%%%%%%%%%%%%%%%%%%%%%%%%%%%%%%%%%%%%%%%%%%%%%%%%%%%%

\section{Methods}

This section describes the existing drift detection methods, such as the Drift Detection Method, the Early Drift Detection Method, the Hoeffding Drift Detection Method, the Kolmogorov-Smirnov test-based Windowing, and Page-Hinkley, as well as our proposed concept drift detection method, Profile Drift Detection.

\subsection{Hoeffding’s Drift Detection Method on Average}
The HDDM method \citep{frias_blanco_2015} includes two different approaches: mean and weighted. Both methods are based on Hoeffding's Inequality. HDDM-A is used to detect abrupt concept drifts in data streams. It monitors changes in the data stream using a simple moving average and detects significant modifications based on Hoeffding's Inequality. Similar to DDM, two different confidence levels are set for alert and concept drift detection. Confidence levels $\alpha_D$ are set for detecting concept drift and $\alpha_W$ for detecting alerts. The total number of samples is denoted by $n$, and the cumulative sum for each new sample is $c = \sum_{i=1}^{n} x_i$. The data stream is divided into two windows: minimum window $(n_{\text{min}}, c_{\text{min}})$ and maximum window $(n_{\text{max}}, c_{\text{max}})$. These windows are used to compare the past and current states of the data. If the minimum and maximum window values are in their initial state, they are updated (if $n_{\text{min}} = 0$ then $n_{\text{min}} = n$, $c_{\text{min}} = c$; if $n_{\text{max}} = 0$ then $n_{\text{max}} = n$, $c_{\text{max}} = c$). Error bounds for the minimum window and total sample size are calculated using Hoeffding's Inequality:
\begin{equation}
\epsilon_{\alpha_D} = \sqrt{\frac{1}{2n} \ln \left(\frac{1}{\alpha_D}\right)}
\end{equation}

\begin{equation}
\epsilon_{\alpha_{D1}} = \sqrt{\frac{1}{2n_{\text{min}}} \ln \left(\frac{1}{\alpha_D}\right)}
\end{equation}

\begin{equation}
\epsilon_{\alpha_{D2}} = \sqrt{\frac{1}{2n_{\text{max}}} \ln \left(\frac{1}{\alpha_D}\right)}
\end{equation}

\noindent Using these bounds, the values of $n_{\text{min}}$, $c_{\text{min}}$, $n_{\text{max}}$, and $c_{\text{max}}$ are updated as follows:

\begin{equation}
\label{eq1}
\text{If } \frac{c_{\text{min}}}{n_{\text{min}}} + \epsilon_{\alpha_{D1}} \geq \frac{c}{n} + \epsilon_{\alpha_D} \text{ then } c_{\text{min}} = c, \; n_{\text{min}} = n
\end{equation}

\begin{equation}
\label{eq2}
\text{If } \frac{c_{\text{max}}}{n_{\text{max}}} - \epsilon_{\alpha_{D2}} \leq \frac{c}{n} - \epsilon_{\alpha_D} \text{ then } c_{\text{max}} = c, \; n_{\text{max}} = n
\end{equation}

\begin{equation}
\frac{c}{n} - \frac{c_{\text{min}}}{n_{\text{min}}} \geq \sqrt{\frac{n - n_{\text{min}}}{n_{\text{min}}} \frac{1}{2n} \ln \left(\frac{1}{\alpha}\right)}
\end{equation}

\begin{equation}
\frac{c_{\text{max}}}{n_{\text{max}}} - \frac{c}{n} \geq \sqrt{\frac{n - n_{\text{max}}}{n_{\text{max}}} \frac{1}{2n} \ln \left(\frac{1}{\alpha}\right)}
\end{equation}

\noindent When the conditions in Equation \ref{eq1} or \ref{eq2} are met, with $\alpha$ replaced by $\alpha_D$, a concept drift is detected, and an alert state is identified when $\alpha_W$ is used.

%%%%%%%%%%%%%%%%%%%%%%%%%%%%%%%%%%%%%%%%%%%%%%%%%%%%%%%%%%%%%%%%%%%%%%%
\subsection{Hoeffding’s Drift Detection Method on Weighted}

HDDM-W is an algorithm that uses Exponentially Weighted Moving Averages (EWMA) to detect concept drifts in data streams. This algorithm tracks average changes occurring in the data stream and determines whether these changes are significant using Hoeffding's Inequality. The parameter \(\lambda\) is used in the EWMA statistic, where smaller values give less weight to the most recent observations and \(\lambda\) ranges from 0 to 1. It is calculated as \(\hat{\mu}_{\text{EWMA}} = \lambda_{x_i} + (1 - \lambda) \hat{\mu}_{\text{EWMA}}\). Similarly, separate \(\alpha\) values must be set for concept drift and alert states.

While HDDM-A is generally more effective in detecting abrupt concept drifts, HDDM-W performs better in detecting gradual concept drifts. HDDM algorithms do not make any distribution assumptions for the observations being compared. Variables can take values in the range \([a, b]\), and Hoeffding's Inequality can be used. However, in practice, Python modules such as scikit-multiflow and river, similar to DDM and EDDM, attempt to detect concept drift based on whether the response variable is correctly predicted.

%%%%%%%%%%%%%%%%%%%%%%%%%%%%%%%%%%%%%%%%%%%%%%%%%%%%%%%%%%%%%%%%%%%%%%%
\subsection{Kolmogorov-Smirnov Windowing Method}
KSWIN is a non-parametric statistical method used to detect concept drifts in data streams \citep{raab_2020}. This method utilizes the Kolmogorov-Smirnov test (KS test) to identify statistical differences between new and historical data. It operates using a sliding window (\(\Psi\)) that continuously holds the most recent \(n\) data points. Two sub-windows are created from \(\Psi\):\\

\noindent \(R\), which contains the most recent \(r\) observations:
\begin{equation}
    R = \{x_i \in \Psi \mid i > n - r\}
\end{equation}

\noindent \(W\), which contains \(r\) observations randomly chosen from the older part of the window:
\begin{equation}
    W = \{x_i \in \Psi \mid i \leq n - r \text{ and } P(x) = \text{UNIF}(x_i \mid 1, n - r)\}
\end{equation}

\noindent where \(|W| = |R| = r\). The Kolmogorov-Smirnov (KS) test is then applied to compare the empirical cumulative distributions of \(R\) and \(W\). The test calculates the maximum absolute difference between these distributions (\(dist_{w,r}\)):

\begin{equation}
dist_{w,r} = \sup_x \left| F_W(x) - F_R(x) \right|
\end{equation}

\noindent If this distance exceeds the critical value (\(D_\alpha\)), it indicates a concept drift:

\begin{equation}
dist_{w,r} > c(\alpha) \sqrt{\frac{n + r}{nr}} = \sqrt{-\frac{1}{2} \ln \alpha} \sqrt{\frac{n + r}{nr}} = \sqrt{-\frac{\ln \alpha}{r}}
\end{equation}

\noindent When the distribution \(P(X)\) changes over time while \(P(y \mid X)\) remains the same, this indicates a virtual concept drift. The KSWIN test is used to detect concept drift under the assumption that any change in \(P(X)\) will also result in a change in \(P(y \mid X)\). Since the KSWIN test is non-parametric, there is no need to assume any specific distribution for the monitored variable. The KSWIN test can be applied to either the response variable or explanatory variables and is used to detect data drift even before actual values are observed, by using explanatory variables to predict the response variable.
%%%%%%%%%%%%%%%%%%%%%%%%%%%%%%%%%%%%%%%%%%%%%%%%%%%%%%%%%%%%%%%%%%%%%%%

\subsection{Page-Hinkley Test}
Page-Hinkley Test (PH) is a method used to detect changes in mean values in data streams, similar to Cumulative Sums \citep{sebastiano_fernandes_2017}. This method is designed to determine whether the average of the data obtained over a certain observation period remains constant. The key difference is the \(\alpha\) forgetting factor, which weighs the observed values and the average, reducing the impact of older data. In the algorithm, initially \(n = 1\), \(x_{\text{ort}} = 0\), and \(S = 0\) are set. For each incoming \(t\)-th observation, the following operations are performed:

\begin{equation}
x_t = x_{\text{ort}} + \frac{x_t - x_{t-1}}{n}
\end{equation}

\begin{equation}
S_t = \max \left(0, \alpha \cdot S_{t-1} + x_t - x - \delta \right)
\end{equation}

\noindent If \(S_t > \lambda\), it indicates the detection of a concept drift. PH can be used to test whether there is a change in the distribution of explanatory variables or the response variable, similar to the KSWIN test. For detecting concept drift, it is not necessary to wait for the actual value of the response variable predicted by the model. The PH test does not include an alert state.
%%%%%%%%%%%%%%%%%%%%%%%%%%%%%%%%%%%%%%%%%%%%%%%%%%%%%%%%%%%%%%%%%%%%%%%

\subsection{Drift Detection Method}

The Drift Detection Method (DDM) is used to detect changes in the response variable (class label) in binary classification problems \citep{gama_2004}. This method is applied in situations where the data stream is continuous and is used to evaluate the model's performance each time a new example arrives. It is assumed that new observations arrive with true values in the form of \((x_i, y_i)\). The model predicts the value \(\hat{y_i}\) for the incoming observations as either Correct or Incorrect. The prediction values being Correct or Incorrect can be modeled with a Bernoulli distribution. When \(n\) observations are received, this distribution is modeled with a Binomial distribution. Let \(i\) be the number of examples:\\

\noindent \textbf{Error Rate} (\(p_i\)): The proportion of errors made by the model at the \(i\)-th example. It is calculated over a specific set of examples and is the ratio of incorrectly classified examples to the total number of examples.\\

\noindent \textbf{Standard Deviation} (\(s_i\)): A value measuring the variance of the error rate at the \(i\)-th example, calculated as \(s_i = \sqrt{p_i(1 - p_i)/i}\).\\

\noindent The error rate (\(p_i\)) is a random variable derived from Bernoulli trials and represented by a binomial distribution. For sufficiently large sample sizes, the binomial distribution approximates a normal distribution with the same mean and variance. The confidence interval for \(p_i\) at \(1 - \alpha/2\) is calculated as \(p_i \pm \alpha \cdot s_i\).

DDM monitors an increase in the error rate. If \((p_i + s_i)\) exceeds a certain threshold, it indicates concept drift. For a 95\% confidence interval, a warning level is suggested as \(p_i + s_i \geq p_{\text{min}} + 2 \cdot s_{\text{min}}\). For a 99\% confidence interval, a drift level is suggested as \(p_i + s_i \geq p_{\text{min}} + 3 \cdot s_{\text{min}}\). DDM is a model-independent, easily applicable algorithm. Since it only attempts to detect concept drift through the response variable, it cannot be used to detect virtual concept drift occurring in explanatory variables.
%%%%%%%%%%%%%%%%%%%%%%%%%%%%%%%%%%%%%%%%%%%%%%%%%%%%%%%%%%%%%%%%%%%%%%%

\subsection{Early Drift Detection Method}

The Early Drift Detection Method (EDDM) algorithm is similar to the DDM algorithm and is designed to detect concept drift in binary classification problems \citep{baena_garcia_2006}. The main difference is that, while DDM calculates the error rate for each incoming \((x_i, y_i)\) observation, EDDM calculates the distance between these errors.

The average distance between two errors \(p'_i\) and the standard deviation of this distance \(s'_i\) are computed and recorded. These values are saved when \(p'_i + 2 \cdot s'_i\) reaches its maximum \((p'_{\text{max}} \text{ and } s'_{\text{max}})\). This maximum value indicates the moment when the model best represents the current concepts in the dataset. EDDM, like DDM, has two different threshold values for warning and drift levels:\\

\textbf{Warning Level:} 
\begin{equation}
  \frac{p'_i + 2 \cdot s'_i}{p'_{\text{max}} + 2 \cdot s'_{\text{max}}} < \alpha
\end{equation}
where \(\alpha\) is suggested to be 0.95.\\
  
\textbf{Drift Level:} 
\begin{equation}
  \frac{p'_i + 2 \cdot s'_i}{p'_{\text{max}} + 2 \cdot s'_{\text{max}}} < \beta
\end{equation}
where \(\beta\) is suggested to be $0.90$.\\

After detecting concept drift, the model is assumed to be retrained to fit the new concept, and \(p'_{\text{max}}\) and \(s'_{\text{max}}\) values are reset. EDDM is designed to detect slowly better-occurring concept drifts and performs well against sudden changes. Like DDM, EDDM detects concept drift based on the response variable, so it cannot be used to detect virtual concept drift occurring in the explanatory variables.
%%%%%%%%%%%%%%%%%%%%%%%%%%%%%%%%%%%%%%%%%%%%%%%%%%%%%%%%%%%%%%%%%%%%%%%

\subsection{Profile Drift Detection}

\begin{figure*}[h]
    \centering
    \includegraphics[width = \linewidth]{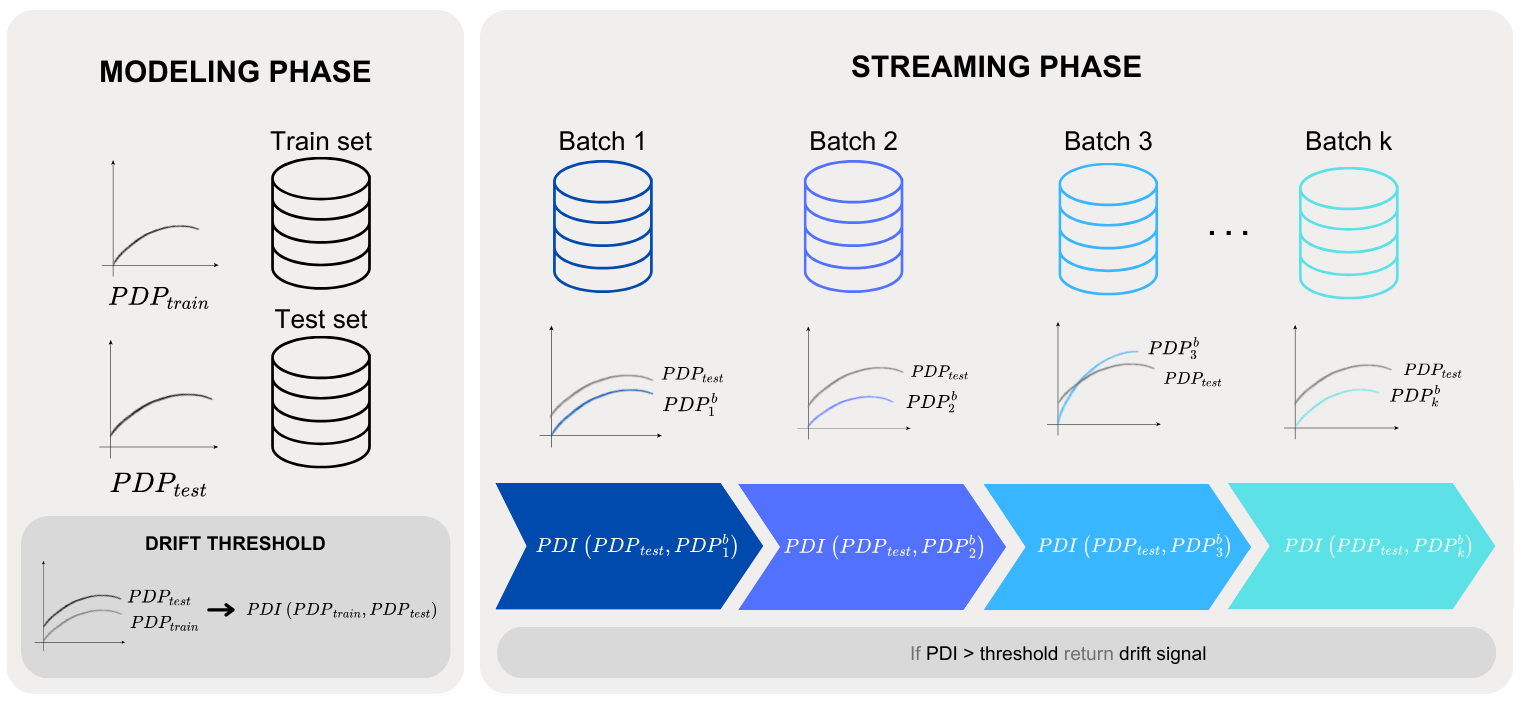}
    \caption{The workflow of the Profile Drift Detection}
    \label{fig:pdd}
\end{figure*}

The combination of L2 distance, first-order derivative distance (L2Der), and the Partial Dependence Index (PDI) methods allows for a comprehensive assessment of various aspects of PDP comparisons \citep{kobylinska_2024}. L2 distance measures the overall difference and magnitude between two profiles. This method evaluates the structural similarity or differences between PDP shapes by considering squared differences between values across all points in the profiles. The L2Der compares the derivatives of profiles to analyze the rate and severity of changes within profiles. Comparing derivatives effectively identifies differences in slopes and local changes, aiding in understanding both the general shapes and dynamic behaviors of profiles. PDI focuses on whether profiles show an increase or decrease at the same data points, evaluating behavioral similarities by taking into account orientation changes between profiles. This index quantitatively expresses differences in profiles’ trends and reveals opposing or similar behaviors. The use of these three methods together provides a holistic measure and analysis of the distance, change rates, and orientation differences between profiles. This allows for a deeper understanding of both the general similarity and behavior at different data points within PDPs. The difference in PDPs analyzed with these three metrics is referred to as PDD, illustrated in Figure~\ref{fig:pdd}.

The three metrics used by PDD (PDI, L2, L2Der) can be set as parameters. In our prior studies \citep{dar_2023,dar_2024}, only the PDI metric was used to attempt to detect concept drift, but it was found to be insufficient on its own in some cases. PDD, on the other hand, incorporates both the L2 and L2Der metrics in addition to PDI. The thresholds set for the three metrics in Algorithm~\ref{algo} should be appropriate for \textit{acceptable} deviations.

\begin{algorithm}[h]
\small
\caption{PDD and Model Updating}
\label{algo}
\begin{algorithmic}[1]
    \State \textbf{Input:} Train and test sets
    \State \textbf{Output:} $test\_accuracy$, $train\_accuracy$, $drift\_count$, $accuracy$
    
    \State Calculate the sizes of the train and test sets according to the given number of batches
    \State $model \gets$ model trained on the train set
    \State $test\_accuracy, train\_accuracy \gets$ Calculate accuracy rates for train and test sets
    \State $variable \gets$ Find the most important variable using permutation importance

    \State $profile1\_train \gets$ Calculate PDP for \textit{variable} on the train set
    \State $profile1 \gets$ Calculate PDP for \textit{variable} on the test set
    \State Create and save the graph using the calculated PDPs
    \State $PDI, L2, L2Der \gets$ Calculate from $profile1\_train$ and $profile1$
    \State $pdi\_threshold \gets PDI$
    \State $l2\_threshold \gets L2$
    \State $l2der\_threshold \gets L2Der$

    \For{each new batch in the dataset from the length of the test set to the length of the dataset, with the batch size as the increment step}
        \State $accuracy\_list \gets$ Calculate the accuracy for the new batch using the model and save it
        \State $profile2 \gets$ Calculate PDP for \textit{variable} on the new batch
        \State $PDI, L2, L2Der \gets$ Calculate metrics from $profile1$ and $profile2$
        
        \If{$PDI > pdi\_threshold$ \textbf{and} $L2 > l2\_threshold$ \textbf{and} $L2Der > l2der\_threshold$}
            \State $drift\_count \gets drift\_count + 1$
            \State Add the data in the new batch on top of the previous data
            \State Train and update the model with the new data
            \State $profile1 \gets$ Calculate PDP for \textit{variable} on the combined data with the new model
            \State Create and save the graph for $profile1$ and $profile2$
        \EndIf
    \EndFor

    \State \Return ($test\_accuracy$, $train\_accuracy$, $drift\_count$, $accuracy = mean(accuracy\_list)$)

\end{algorithmic}
\end{algorithm}

In Algorithm~\ref{algo}, three metrics calculated over train-test data are compared with metrics obtained from the new stream. A large value for all three metrics indicates concept drift. However, depending on the structure of the dataset, the new data (and hence the PDPs) may be similarly shaped but separated along the vertical axis, which could lead to false negatives if the PDI metric is used. High metric values in the train and test sets can produce false positives in the new batches. Similarly, statistical tests have different sensitivity parameters. For example, the KSWIN test has parameters such as \(\alpha\), window size, and size. Similarly, PH, HDDM-W, HDDM-A, EDDM, and DDM have three parameters each. Changing the parameters of PDD and statistical tests affects the sensitivity of drift detection. 
%%%%%%%%%%%%%%%%%%%%%%%%%%%%%%%%%%%%%%%%%%%%%%%%%%%%%%%%%%%%%%%%%%%%%%%%%%%%%

\section{Experiments} 

This section compares traditional methods and the PDD method using synthetic and real-world datasets. It also describes the characteristics of the benchmark dataset and the design of the experiment.

%%%%%%%%%%%%%%%%%%%%%%%%%%%%%%%%%%%%%%%%%%%%%%%%%%%%%%%%%%%%%%%%%%%%%%%
\subsection{Datasets}

In the experiments, we used the generic synthetic and real-world datasets \texttt{SEA} \citep{street_and_kim_2001}, \texttt{Hyperplane} \citep{hulten_et_al_2001}, \texttt{NOAA} \citep{elwell_polikar_2011}, \texttt{Ozone} \citep{zhang_and_fan_2008}, \texttt{Elec2} \citep{harries_1999}, and \texttt{Friedman} \citep{ikonomovska_et_al_2011}. The characteristics of these datasets are provided in Table~\ref{tab:dataset}.

\begin{table}[h]
    \centering
    \scriptsize
    \caption{The characteristics of the benchmark dataset}
    \label{tab:dataset}
    \begin{tabular}{lllrr}\toprule
    \textbf{Dataset}        & \textbf{Type} & \textbf{Task}     & \textbf{\#observations}   & \textbf{\#variables} \\\midrule
    \texttt{SEA}            & Synthetic     & Classification    & 50000            & 4\\
    \texttt{Hyperplane}     & Synthetic     & Classification    & 20000            & 3\\
    \texttt{NOAA}           & Real          & Classification    & 18159            & 9\\
    \texttt{Ozone}          & Real          & Classification    & 2534             & 73\\
    \texttt{Elec2}          & Real          & Classification    & 45312            & 9\\
    \texttt{Friedman}       & Synthetic     & Regression        & 20000            & 11\\\bottomrule
    \end{tabular}
\end{table}

%%%%%%%%%%%%%%%%%%%%%%%%%%%%%%%%%%%%%%%%%%%%%%%%%%%%%%%%%%%%%%%%%%%%%%%
\subsection{Design of experiments}

In the experiments, Logistic Regression (LR), Decision Tree (DT), and Random Forest (RF) models were used. However, for the Friedman dataset, linear regression was used instead of Logistic Regression.

For incoming data batches, it is determined whether the data is suitable for training the model. PDI, L2Der, and L2 are calculated from PDPs obtained from the train and test sets and are assigned as threshold values. If these three metrics exceed the specified thresholds, the new data batch is added to the train set, and the model is retrained. For the batches where the PDD method detects concept drift, PDP curves for the train-test sets and PDP graphs at the time of drift detection are included for each dataset. This process helps in detecting concept drift and allows the model to adapt to new data. Thus, the model is continuously updated and can adapt to changes in the data stream. 

Statistical tests are conducted to verify whether the response variable in the incoming data batch is predicted correctly. For each new data batch, predicted values are obtained and tested against the actual values. If concept drift is detected in this data batch, similar incoming data is added to the train data, and the model is retrained.  

For each experiment, Tables~\ref{tab:exp_SEA}-\ref{tab:exp_Friedman} show the concept drift detection method used, the number of batches, the size of the test set (equal to the batch size), the size of the training set, the number of detected drifts for each batch, and the model's performance on the training and testing sets. A larger batch size may lead to missed detections of drifts, while a smaller size can result in false detections. Therefore, determining the optimal batch size (or number of batches) is essential.

Additional information about the experiments, such as the number of observations in the batches (Table~\ref{tab:batch_number}), the most important variables in the batches (Table~\ref{tab:vip_batch}), and the accuracy of the models in the batches (Table~\ref{tab:acc_batch}) are given in the Appendix.

%%%%%%%%%%%%%%%%%%%%%%%%%%%%%%%%%%%%%%%%%%%%%%%%%%%%%%%%%%%%%%%%%%%%%%%
\subsection{Results}

In this section, the results of experiments conducted on the datasets \texttt{SEA}, \texttt{Hyperplane}, \texttt{NOAA}, \texttt{Ozone}, \texttt{Elec2}, and \texttt{Friedman} are summarized. The performance of the methods is evaluated in terms of accuracy and the number of detected drifts (\texttt{\#drift}). The PDI metric was not used due to the similarity in shape, and the cutoff values for L2 and L2Der were set to five times the values calculated on the test set in the logistic regression models. All metrics were initialized on the test sets and used as cutoff values.

%%%%%%%%%%%%%%%%%%%%%%%%%%%%%%%%%%%%%%%%%%%%%%%%%%%%%%%%%%%%%%%%%%%%%%%
\subsubsection{\texttt{SEA}}

Table~\ref{tab:exp_SEA} presents the performance of various drift detection methods applied to three models (LR, DT, RF) over different batch sizes. Overall, accuracy values remain quite similar across methods, generally ranging between 0.82 and 0.85. However, there are notable differences in how many drift events each method detects.

% SEA 
\begin{table}[h]
    \centering
    \scriptsize
    \caption{The results of experiments on the \texttt{SEA} dataset}
    \label{tab:exp_SEA}
    \resizebox{\linewidth}{!}{
    \begin{tabular}{llcccccc}\toprule
        && \multicolumn{6}{c}{\textbf{\#batch}}\\\cmidrule(lr){3-8}
        && \multicolumn{2}{c}{\textbf{10}}    & \multicolumn{2}{c}{\textbf{20}}    & \multicolumn{2}{c}{\textbf{30}} \\\cmidrule(lr){3-8}
        \textbf{Method} & \textbf{Model} & \textbf{Acc.} & \textbf{\#drifts} & \textbf{Acc.} & \textbf{\#drifts} & \textbf{Acc.} & \textbf{\#drifts} \\\midrule
        % HDDM-A
        HDDM-A & LR   & 0.8306 & 1  & 0.8462 & 2  & 0.8446 & 2 \\
               & DT   & 0.8318 & 2  & 0.8399 & 2  & 0.8357 & 2 \\
               & RF   & 0.8306 & 1  & 0.8416 & 2  & 0.8382 & 2 \\\midrule
        % HDDM-W
        HDDM-W & LR   & 0.8279 & 3  & 0.8442 & 2  & 0.8407 & 2 \\
               & DT   & 0.8247 & 3  & 0.8389 & 2  & 0.8354 & 2 \\
               & RF   & 0.8258 & 3  & 0.8411 & 4  & 0.8370 & 2 \\\midrule
        % KSWIN
        KSWIN  & LR   & 0.8292 & 10 & 0.8454 & 20 & 0.8422 & 30 \\
               & DT   & 0.8293 & 10 & 0.8440 & 20 & 0.8376 & 29 \\
               & RF   & 0.8267 & 10 & 0.8421 & 20 & 0.8395 & 28 \\\midrule
        % PageHinkley (PH)
        PH     & LR   & 0.8337 & 1  & 0.8446 & 3  & 0.8437 & 3 \\
               & DT   & 0.8307 & 1  & 0.8424 & 3  & 0.8367 & 3 \\
               & RF   & 0.8300 & 1  & 0.8417 & 3  & 0.8391 & 3 \\\midrule
        % DDM
        DDM    & LR   & 0.8283 & 0  & 0.8459 & 3  & 0.8426 & 5 \\
               & DT   & 0.8267 & 0  & 0.8432 & 1  & 0.8356 & 3 \\
               & RF   & 0.8268 & 0  & 0.8427 & 3  & 0.8377 & 3 \\\midrule
        % EDDM
        EDDM   & LR   & 0.8292 & 10 & 0.8454 & 20 & 0.8381 & 12 \\
               & DT   & 0.8290 & 9  & 0.8435 & 18 & 0.8340 & 8 \\
               & RF   & 0.8267 & 10 & 0.8421 & 19 & 0.8391 & 13 \\\midrule
        % PDD
        PDD    & LR   & 0.8283 & 0  & 0.8352 & 0  & 0.8391 & 0 \\
               & DT   & 0.8230 & 1  & 0.8324 & 3  & 0.8349 & 8 \\
               & RF   & 0.8283 & 5  & 0.8348 & 8  & 0.8376 & 18 \\\bottomrule
    \end{tabular}}
\end{table}

For the LR model, methods such as KSWIN and EDDM register a very high number of drift detections (10, 20, and 30 or 10, 20, and 12 for 10, 20, and 30 batches, respectively). In contrast, the PDD method did not detect any drifts at any batch size, and other methods like PH and DDM reported minimal detections (e.g., PH: 1, 3, 3; DDM: 0, 3, 5). This suggests that, in the LR model, the underlying Partial Dependence Profiles (PDPs) might be highly consistent between training and test sets—resulting in a near-zero Profile Disparity Index (PDI)—which in turn prevents the PDD method from flagging any drift even when other methods signal significant changes.

While accuracy remains comparable across methods in the DT model, drift counts differ. HDDM-A and HDDM-W consistently detect 2 or 3 drifts, whereas KSWIN and EDDM report much higher counts (around 10 to 29 drifts across different batch sizes). The PDD method, on the other hand, starts with a low detection count (1 drift at 10 batches) but increases to 8 drifts by the 30th batch, indicating a moderate sensitivity that grows with batch size.

For the RF model, a similar pattern emerges. The PDD method detects 5 drifts at 10 batches, increasing to 8 and then 18 drifts as the batch size grows. This is intermediate compared to KSWIN and EDDM (which report 10–28 and 10–19–13 drifts, respectively) and the more conservative methods like DDM and PH (which remain within 0–3 drifts).

In summary, while all methods maintain similar accuracy levels, there is a clear trade-off in drift detection sensitivity. Methods like KSWIN and EDDM are very aggressive in flagging drift events, potentially increasing the risk of false positives, especially in operational settings. Conversely, the PDD method exhibits a more conservative behavior—particularly in the LR model, where it detects no drifts—suggesting that it might miss subtle changes when the PDPs remain nearly parallel. For DT and RF models, PDD’s detection rate increases with larger batch sizes, but it remains more controlled compared to the more sensitive methods. This variability underscores the importance of choosing a drift detection strategy that aligns with both the model characteristics and the operational requirements of the system.

\begin{figure}[h]
    \centering
    \includegraphics[width = \linewidth]{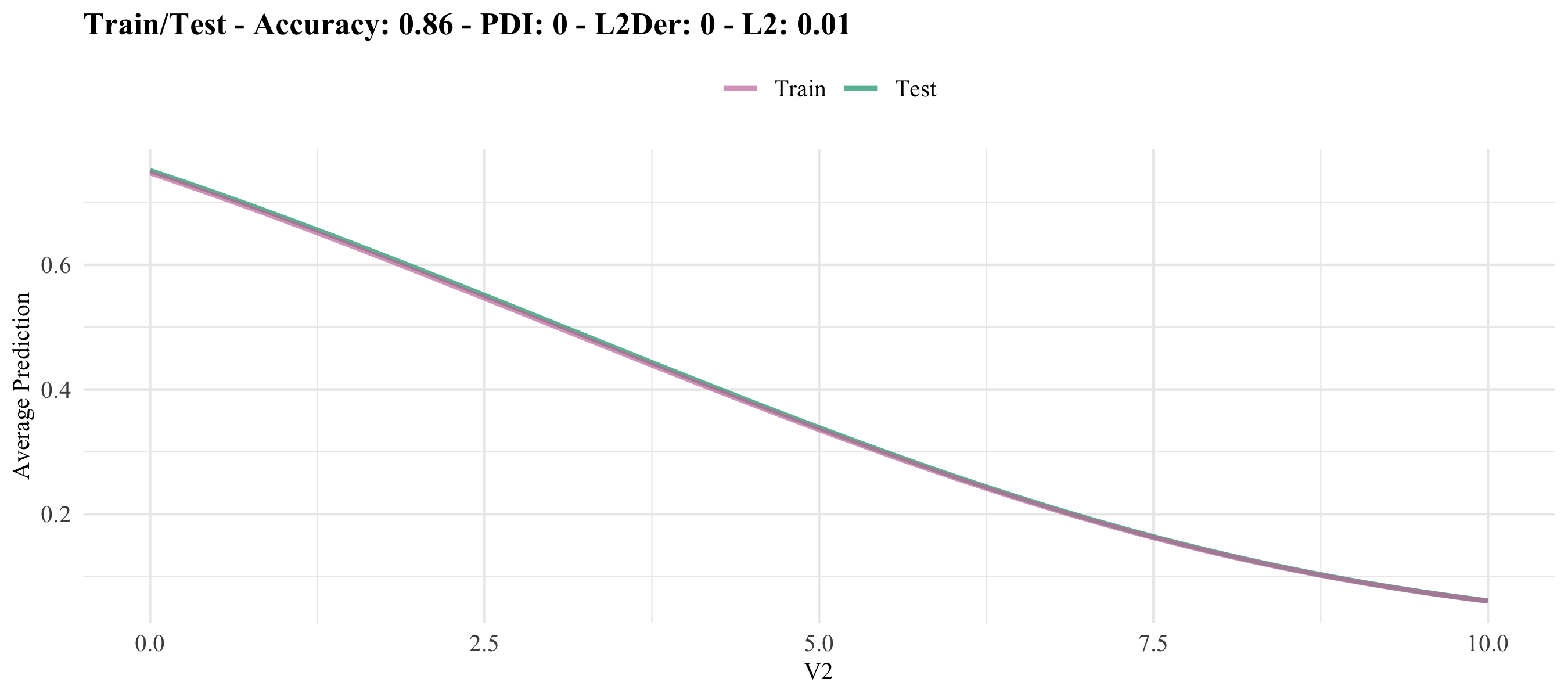}
    \caption{PDPs of the Logistic Regression model trained on \texttt{SEA} dataset}
    \label{fig:sealogistic}
\end{figure}

In the LR model, no drift was detected—even during the 10th batch. As illustrated in Figure~\ref{fig:sealogistic}, the PDPs for the train and test sets overlap almost perfectly, indicating that the functional relationship between the input variables and the response has remained stable. This complete overlap results in a PDI of 0, demonstrating that despite any shifts in data distribution, the model’s predictive behavior has not been affected.

\begin{figure}[h]
    \centering
    \includegraphics[width = \linewidth]{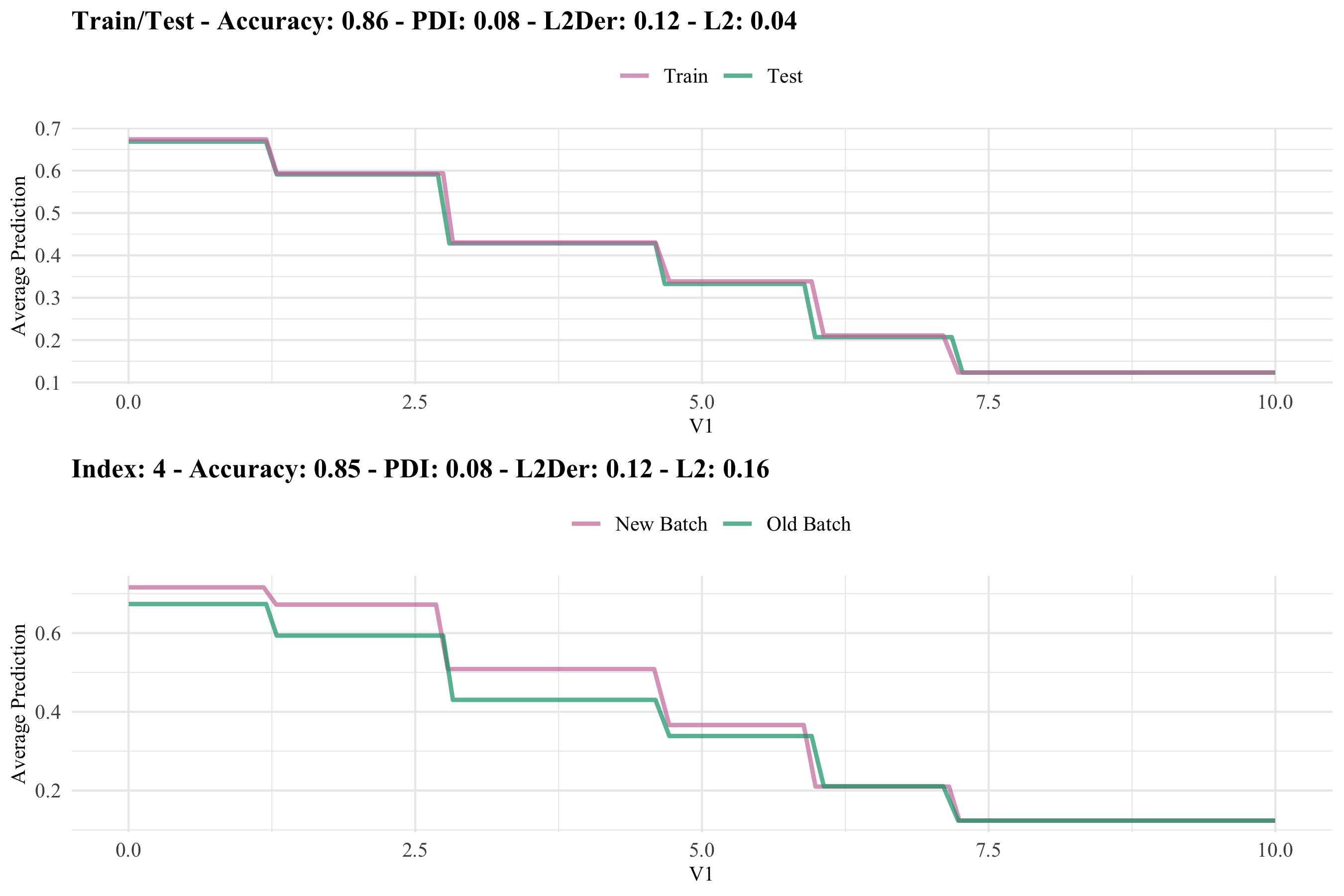}
    \caption{PDPs of the Decision Tree model trained on \texttt{SEA} dataset}
    \label{fig:seadecisiontree}
\end{figure}

For the DT model with 10 batches, drift was detected in the 4th batch. As shown in Figure~\ref{fig:seadecisiontree}, it is evident that the variable \texttt{V1} plays a significant role in this drift. Specifically, within the range of 0 to 5, the contribution of \texttt{V1} to the average response variable shows a noticeable decline. This suggests that within this particular range, the predictive influence of \texttt{V1} weakens, which likely contributes to the observed drift. The reduction in \texttt{V1}'s contribution implies that its relationship with the target variable has shifted in this batch, potentially due to underlying changes in the data distribution. Such a shift highlights the importance of monitoring variable contributions closely over time, as even subtle alterations can signal data drifts that may ultimately affect model performance.

\begin{figure}[h]
    \centering
    \includegraphics[width = \linewidth]{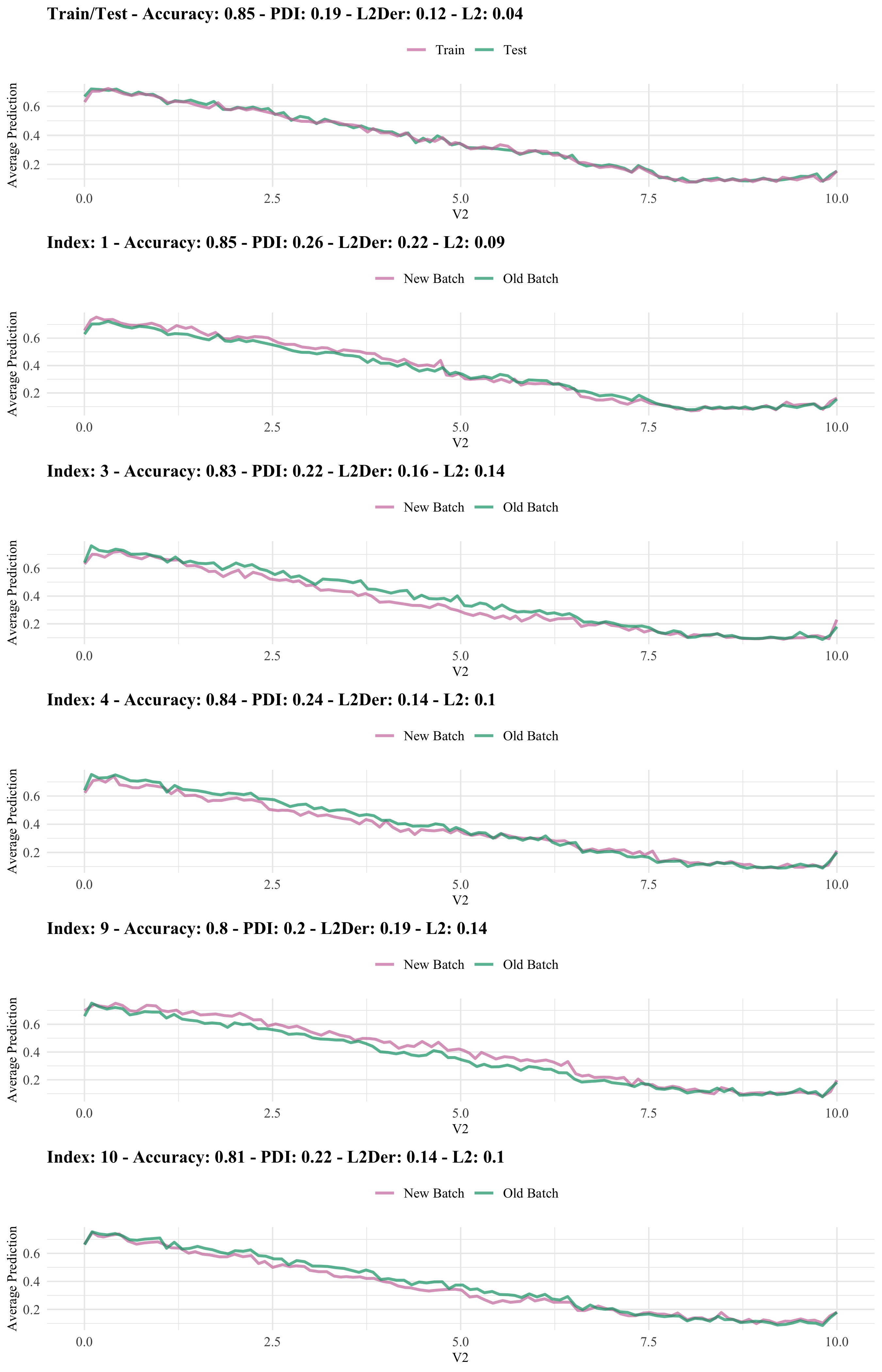}
    \caption{PDPs of the Random Forest model trained on \texttt{SEA} dataset}
    \label{fig:searandomforest}
\end{figure}

The variable \texttt{V2} has been identified as the most important variable in the RF model. Figure~\ref{fig:searandomforest} shows that \texttt{V2} is the most important variable in the model’s predictions. Specifically, within the range of 0 to 7.5, noticeable variations in the behavior of \texttt{V2} are evident. These changes suggest that the influence of \texttt{V2} on the model's output has shifted in these specific instances, potentially due to variations in the data distribution during these batches. The observed changes in \texttt{V2} highlight how shifts in important variables can significantly impact the model's performance and its ability to generalize well to incoming data. This further emphasizes the importance of monitoring key variables closely, as even small changes in their behavior can lead to drift detection and the need for model adaptation, particularly when dealing with real-time data streams.

%%%%%%%%%%%%%%%%%%%%%%%%%%%%%%%%%%%%%%%%%%%%%%%%%%%%%%%%%%%%%%%%%%%%%%%
\subsubsection{\texttt{Hyperplane}}

The results in Table~\ref{tab:exp_Hyperplane} for the \texttt{Hyperplane} dataset reveal distinct performance trends across drift detection methods and models, with the PDD method showing mixed outcomes.

% Hyperplane
\begin{table}[h]
    \centering
    \scriptsize
    \caption{The results of experiments on \texttt{Hyperplane} dataset}
    \label{tab:exp_Hyperplane}
    \resizebox{\linewidth}{!}{
    \begin{tabular}{llcccccc}\toprule
       && \multicolumn{6}{c}{\textbf{\#batch}}\\\cmidrule(lr){3-8}
       && \multicolumn{2}{c}{\textbf{10}} & \multicolumn{2}{c}{\textbf{20}} & \multicolumn{2}{c}{\textbf{30}} \\\cmidrule(lr){3-8}
       \textbf{Method} & \textbf{Model} & \textbf{Acc.} & \textbf{\#drifts} & \textbf{Acc.} & \textbf{\#drifts} & \textbf{Acc.} & \textbf{\#drifts} \\\midrule
       % HDDM-A
       HDDM-A & LR & 0.8662 & 1 & 0.8204 & 2 & 0.7568 & 3 \\
              & DT & 0.8398 & 0 & 0.7670 & 3 & 0.6921 & 3 \\
              & RF & 0.7834 & 0 & 0.6809 & 1 & 0.6267 & 1 \\\midrule
       % HDDM-W
       HDDM-W & LR & 0.8377 & 1 & 0.8456 & 3 & 0.8152 & 4 \\
              & DT & 0.8398 & 0 & 0.7752 & 2 & 0.7316 & 4 \\
              & RF & 0.7794 & 2 & 0.7615 & 3 & 0.7536 & 6 \\\midrule
       % KSWIN
       KSWIN  & LR & 0.8639 & 9 & 0.8458 & 13 & 0.8285 & 18 \\
              & DT & 0.8520 & 9 & 0.8429 & 17 & 0.8162 & 19 \\
              & RF & 0.8219 & 10 & 0.8144 & 17 & 0.7948 & 22 \\\midrule
       % PageHinkley
       PH & LR & 0.8648 & 0 & 0.8187 & 1 & 0.7455 & 2 \\
                   & DT & 0.8398 & 0 & 0.7014 & 1 & 0.6929 & 1 \\
                   & RF & 0.7834 & 0 & 0.7033 & 1 & 0.6267 & 1 \\\midrule
       % DDM
       DDM    & LR & 0.8648 & 0 & 0.8071 & 1 & 0.7455 & 2 \\
              & DT & 0.8398 & 0 & 0.7014 & 1 & 0.7283 & 3 \\
              & RF & 0.7834 & 0 & 0.7290 & 3 & 0.6267 & 1 \\\midrule
       % EDDM
       EDDM   & LR & 0.8648 & 0 & 0.8485 & 13 & 0.8348 & 5 \\
              & DT & 0.8496 & 10 & 0.8562 & 15 & 0.8379 & 11 \\
              & RF & 0.7834 & 0 & 0.8072 & 16 & 0.7957 & 24 \\\midrule
       % PDD
       PDD    & LR & 0.8648 & 0 & 0.6792 & 0 & 0.5219 & 0 \\
              & DT & 0.8435 & 1 & 0.8242 & 15 & 0.4924 & 1 \\
              & RF & 0.7831 & 1 & 0.7189 & 4 & 0.5099 & 0 \\\bottomrule
    \end{tabular}}
\end{table}

An examination of the train and test set accuracies reveals a potential overfitting issue, Table~\ref{tab:acc_batch}. For the LR model, most methods yield a high accuracy of approximately 0.86–0.87 at 10 batches. However, while methods such as HDDM-A, HDDM-W, KSWIN, PH, DDM, and EDDM maintain relatively high accuracy at 20 and 30 batches (ranging from about 0.80 to 0.83 at 20 batches and 0.75 to 0.83 at 30 batches), the PDD method experiences a dramatic drop—from 0.8648 at 10 batches to 0.6792 at 20 batches and further to 0.5219 at 30 batches. Notably, PDD does not detect any drift events in the LR model across all batch sizes, which may indicate that it is overly conservative or insensitive to drifts when the model’s predictive behavior is stable, even if the overall performance degrades with larger batch sizes.

In the case of the DT model, accuracy levels at 10 batches are fairly consistent across methods (around 0.84), yet divergence emerges as the number of batches increases. While methods like KSWIN and EDDM continue to yield high accuracies (up to 0.8429 and 0.8562, respectively, at 20 batches, and 0.8162 and 0.8379 at 30 batches), PDD shows an unusual pattern—detecting only 1 drift at 10 and 30 batches but spiking to 15 detected drifts at 20 batches—with corresponding accuracy falling to 0.8242 at 20 batches and dropping to 0.4924 at 30 batches. This inconsistency in drift detection for DT models suggests that PDD’s sensitivity may be erratic, potentially linked to changes in the data stream or model behavior as more batches are processed.

For the RF model, all methods report lower accuracies compared to the LR and DT models. While KSWIN and EDDM achieve relatively high accuracies (0.8144 and 0.8072 at 20 batches, and 0.7948 and 0.7957 at 30 batches, respectively), PDD registers moderate accuracy at 10 batches (0.7831) but then declines to 0.7189 at 20 batches and further plummets to 0.5099 at 30 batches. In terms of drift detection, PDD detects only 1 drift at 10 batches and 4 at 20 batches, but it fails to detect any drifts at 30 batches—contrasting sharply with KSWIN, which detects 10, 17, and 22 drifts across the three batch sizes, and EDDM, which records 0, 16, and 24 drifts. This pattern in the RF model indicates that while PDD’s conservative detection strategy may help limit false alarms, it might also miss genuine drift events, especially as the model’s accuracy declines.

Overall, methods like KSWIN and EDDM consistently register a higher number of drift detections, underscoring their sensitivity to changes in the data stream; however, this heightened sensitivity could potentially lead to more false positives. In contrast, HDDM-A and HDDM-W tend to detect fewer drifts, and methods such as PH and DDM offer moderate performance in both accuracy and drift detection. The PDD method, in particular, demonstrates a mixed profile: it preserves high accuracy in the LR model at 10 batches but suffers a pronounced accuracy decline and fails to detect any drifts as the batch size increases. For DT and RF models, PDD’s drift detection is inconsistent and generally lower compared to its counterparts, with a notable drop in accuracy at higher batch sizes.

These findings suggest that the effectiveness of the PDD method in the Hyperplane dataset is highly dependent on both the model type and the batch size. While its conservative nature may reduce the incidence of false positive drift detections, it also raises concerns about its ability to capture actual drifts when the model’s performance is already compromised. Consequently, the utility of PDD may be limited in scenarios where models are prone to overfitting or where accuracy deteriorates with increasing data volume.

%%%%%%%%%%%%%%%%%%%%%%%%%%%%%%%%%%%%%%%%%%%%%%%%%%%%%%%%%%%%%%%%%%%%%%%
\subsubsection{\texttt{NOAA}}

The results for the \texttt{NOAA} dataset in Table~\ref{tab:exp_NOAA} show that while overall accuracy levels across methods remain relatively stable—with LR and RF models achieving accuracies in the mid- to high 0.70 and DT models around 0.73–0.74, the number of drift detections varies considerably between methods and models.

% NOAA
\begin{table}[h]
    \centering
    \scriptsize
    \caption{The results of experiments on the \texttt{NOAA} dataset}
    \label{tab:exp_NOAA}
    \resizebox{\linewidth}{!}{
    \begin{tabular}{llcccccc}\toprule
       && \multicolumn{6}{c}{\textbf{\#batch}}\\\cmidrule(lr){3-8}
       && \multicolumn{2}{c}{\textbf{10}} & \multicolumn{2}{c}{\textbf{20}} & \multicolumn{2}{c}{\textbf{30}} \\\cmidrule(lr){3-8}
       \textbf{Method} & \textbf{Model} & \textbf{Acc.} & \textbf{\#drifts} & \textbf{Acc.} & \textbf{\#drifts} & \textbf{Acc.} & \textbf{\#drifts} \\\midrule
       % HDDM-A
       HDDM-A & LR & 0.7736 & 5 & 0.7696 & 7 & 0.7777 & 3 \\
              & DT & 0.7293 & 8 & 0.7435 & 9 & 0.7440 & 8 \\
              & RF & 0.7796 & 0 & 0.7812 & 5 & 0.7770 & 2 \\\midrule
       % HDDM-W
       HDDM-W & LR & 0.7736 & 7 & 0.7712 & 12 & 0.7799 & 12 \\
              & DT & 0.7385 & 10 & 0.7345 & 16 & 0.7385 & 23 \\
              & RF & 0.7877 & 4 & 0.7805 & 11 & 0.7824 & 9 \\\midrule
       % KSWIN
       KSWIN  & LR & 0.7755 & 10 & 0.7755 & 20 & 0.7788 & 27 \\
              & DT & 0.7385 & 10 & 0.7334 & 20 & 0.7406 & 30 \\
              & RF & 0.7914 & 10 & 0.7894 & 18 & 0.7922 & 26 \\\midrule
       % PageHinkley
       PH & LR & 0.7707 & 3 & 0.7680 & 3 & 0.7792 & 3 \\
                   & DT & 0.7402 & 2 & 0.7387 & 2 & 0.7276 & 2 \\
                   & RF & 0.7877 & 3 & 0.7784 & 3 & 0.7792 & 3 \\\midrule
       % DDM
       DDM    & LR & 0.7716 & 4 & 0.7746 & 10 & 0.7715 & 10 \\
              & DT & 0.7422 & 3 & 0.7334 & 20 & 0.7398 & 21 \\
              & RF & 0.7809 & 1 & 0.7916 & 20 & 0.7829 & 8 \\\midrule
       % EDDM
       EDDM   & LR & 0.7755 & 10 & 0.7755 & 20 & 0.7780 & 30 \\
              & DT & 0.7385 & 10 & 0.7334 & 20 & 0.7406 & 30 \\
              & RF & 0.7914 & 10 & 0.7916 & 20 & 0.7941 & 30 \\\midrule
       % PDD
       PDD    & LR & 0.7700 & 1 & 0.7694 & 2 & 0.7705 & 4 \\
              & DT & 0.7264 & 2 & 0.7403 & 0 & 0.7352 & 3 \\
              & RF & 0.7881 & 5 & 0.7676 & 0 & 0.7913 & 22 \\\bottomrule
    \end{tabular}}
\end{table}

For the LR model, the PDD method maintains a consistent accuracy of approximately 0.770 across all batch sizes. However, it flags very few drift events, with only 1 drift at 10 batches, 2 at 20 batches, and 4 at 30 batches. In contrast, more sensitive methods such as KSWIN and EDDM detect between 10 and 27 drifts (KSWIN) or 10 and 30 drifts (EDDM) across the same batch sizes.

The DT model exhibits a similar trend in terms of accuracy, where PDD detects only 2 drifts at 10 batches, none at 20 batches, and 3 at 30 batches—again, far fewer than the consistently higher counts produced by KSWIN and EDDM (which report 10, 20, and 30 drifts, respectively).

The RF model presents a different pattern. Although its overall accuracy stays in the 0.78–0.79 range, PDD's drift detection is notably inconsistent: it detects 5 drifts at 10 batches, 0 drifts at 20 batches, and then jumps to 22 drifts at 30 batches. This inconsistency is particularly striking when compared to KSWIN (which detects 10, 18, and 26 drifts) and EDDM (with 10, 20, and 30 drifts). Given that the RF model is known to overfit—evidenced by an almost perfect training accuracy contrasted with its moderate test accuracy—this erratic behavior in PDD’s drift detection may be influenced by the overfitting, which can distort the Partial Dependence Profile (PDP) comparisons used by PDD.

Other methods show their characteristics: PH remains steady with 3 drifts for LR and RF (and 2 for DT) across all batch sizes, while HDDM-A and HDDM-W detect a moderate number of drifts that vary with both model and batch size.

In summary, while the PDD method yields stable accuracy and a controlled number of drift detections for the LR and DT models, its performance in the RF model is inconsistent—likely due to overfitting effects. This suggests that PDD might be less reliable in environments where models are prone to overfitting, as it can under-detect or inconsistently detect drift events compared to more sensitive methods like KSWIN and EDDM, which, although potentially generating more false positives, offer a more consistent detection framework across different batch sizes and models.

%%%%%%%%%%%%%%%%%%%%%%%%%%%%%%%%%%%%%%%%%%%%%%%%%%%%%%%%%%%%%%%%%%%%%%%
\subsubsection{\texttt{Ozone}}

The experimental results for the \texttt{Ozone} dataset, as presented in Table~\ref{tab:exp_Ozone}, offer a detailed picture of both drift detection performance and variable importance dynamics across different models and batch sizes. Overall, accuracy remains high for all models, yet the number of drift detections varies noticeably depending on the method employed. For example, in the LR model, accuracy starts at approximately 0.93 for 10 batches, declines slightly to around 0.89 for 20 batches, and then reaches about 0.87 for 30 batches. At the same time, while methods like HDDM-A, HDDM-W, and PH report minimal drift events—often zero or just a couple—the more sensitive techniques such as KSWIN, DDM, and EDDM tend to detect more drifts, with counts ranging from 3 to 6. The PDD method, in particular, shows moderate sensitivity in the LR model by detecting 2 drifts for both 10 and 20 batches and 3 drifts for 30 batches.

% Ozone
\begin{table}[h]
    \centering
    \scriptsize
    \caption{The results of experiments on the \texttt{Ozone} dataset}
    \label{tab:exp_Ozone}
    \resizebox{\linewidth}{!}{
    \begin{tabular}{llcccccc}\toprule
        && \multicolumn{6}{c}{\textbf{\#batch}}\\\cmidrule(lr){3-8}
        && \multicolumn{2}{c}{\textbf{10}} & \multicolumn{2}{c}{\textbf{20}} & \multicolumn{2}{c}{\textbf{30}} \\\cmidrule(lr){3-8}
        \textbf{Method} & \textbf{Model} & \textbf{Acc.} & \textbf{\#drifts} & \textbf{Acc.} & \textbf{\#drifts} & \textbf{Acc.} & \textbf{\#drifts} \\\midrule
        % HDDM-A
        HDDM-A & LR & 0.9294 & 0  & 0.8942 & 2  & 0.8116 & 0 \\
               & DT & 0.9487 & 0  & 0.9101 & 1  & 0.9036 & 2 \\
               & RF & 0.9508 & 0  & 0.9427 & 1  & 0.9374 & 0 \\\midrule
        % HDDM-W
        HDDM-W & LR & 0.9294 & 0  & 0.8907 & 1  & 0.8526 & 1 \\
               & DT & 0.9487 & 0  & 0.9101 & 1  & 0.9122 & 2 \\
               & RF & 0.9508 & 0  & 0.9427 & 1  & 0.9401 & 1 \\\midrule
        % KSWIN
        KSWIN  & LR & 0.9396 & 3  & 0.9164 & 2  & 0.9144 & 5 \\
               & DT & 0.9444 & 3  & 0.9242 & 4  & 0.9099 & 5 \\
               & RF & 0.9529 & 1  & 0.9383 & 4  & 0.9437 & 3 \\\midrule
        % PageHinkley (PH)
        PH     & LR & 0.9294 & 0  & 0.8502 & 0  & 0.8116 & 0 \\
               & DT & 0.9487 & 0  & 0.8911 & 0  & 0.8725 & 0 \\
               & RF & 0.9508 & 0  & 0.9422 & 0  & 0.9374 & 0 \\\midrule
        % DDM
        DDM    & LR & 0.9406 & 4  & 0.9155 & 6  & 0.8706 & 2 \\
               & DT & 0.9358 & 7  & 0.9301 & 6  & 0.9329 & 4 \\
               & RF & 0.9513 & 2  & 0.9427 & 2  & 0.9437 & 3 \\\midrule
        % EDDM
        EDDM   & LR & 0.9396 & 2  & 0.9183 & 4  & 0.8842 & 6 \\
               & DT & 0.9412 & 2  & 0.9164 & 4  & 0.9185 & 5 \\
               & RF & 0.9524 & 2  & 0.9441 & 3  & 0.9437 & 3 \\\midrule
        % PDD
        PDD    & LR & 0.9300 & 2  & 0.9126 & 2  & 0.8722 & 3 \\
               & DT & 0.9435 & 0  & 0.8915 & 1  & 0.9140 & 10 \\
               & RF & 0.9459 & 5  & 0.9422 & 0  & 0.9455 & 12 \\\bottomrule
    \end{tabular}}
\end{table}

In the DT model, accuracies are consistently high, hovering around 0.94 at 10 batches, with slight variations as batch size increases. While some methods register only a few drift events, the PDD method exhibits an interesting pattern: it does not detect any drift at 10 batches, detects a single drift at 20 batches, and then experiences a sudden increase to 10 drift detections at 30 batches. This abrupt change may be reflective of an evolving decision boundary as the model processes more data. The RF model, on the other hand, maintains high accuracy—often exceeding 0.94—but shows inconsistent drift detection with PDD, which detects 5 drifts at 10 batches, none at 20 batches, and then jumps to 12 drifts at 30 batches. Such variability in the RF model might be linked to its inherent complexity or even issues like overfitting, which can affect how subtle changes in the model’s predictive profile are interpreted by drift detection methods.

Adding another layer to this analysis, the variable importance table (Table~\ref{tab:vip_batch}) provides insights into how the most critical variables evolve with batch size across different datasets and models. In the Ozone dataset, for instance, the RF model shifts its focus from variable \texttt{V56} at 10 batches to \texttt{V61} at 20 and 30 batches, while the LR model moves from \texttt{V25} to \texttt{V19}, and the DT model transitions from \texttt{V31} to \texttt{V36} as the batch size increases. These changes suggest that as more data is incorporated, the models may be re-weighting the influence of certain variables, which could lead to alterations in the shape of the PDPs and, consequently, affect drift detection outcomes. 

When considering the interplay between drift detection and variable importance, it becomes clear that methods that rely on comparing the behavior of key variables, such as the PDD approach, might register drift events not solely due to shifts in data distribution but also due to changes in the relative importance of those variables. The sudden increase in drift detections by PDD in the DT model at 30 batches could be partly attributed to its change in the most influential variable from \texttt{V31} to \texttt{V36}. Similarly, the inconsistent detection pattern observed in the RF model with PDD might be related to how the model’s primary variables are adjusted as more batches are processed.

In summary, the results indicate that while high accuracy is maintained across models in the Ozone dataset, the sensitivity of drift detection methods varies significantly. Conservative methods tend to produce fewer detections, whereas more sensitive ones capture a higher number of drift events, which may lead to false positives. The PDD method, although demonstrating moderate performance in the LR model, exhibits inconsistent behavior in the DT and RF models—particularly as batch size increases—potentially due to changes in variable importance or model overfitting. This analysis underscores the complexity of drift detection in dynamic environments and highlights the importance of monitoring both model accuracy and feature relevance over time to refine drift detection strategies effectively.

%%%%%%%%%%%%%%%%%%%%%%%%%%%%%%%%%%%%%%%%%%%%%%%%%%%%%%%%%%%%%%%%%%%%%%%
\subsubsection{\texttt{Elec2}}

The results for the \texttt{Elec2} dataset reveal distinct performance patterns across drift detection methods, models, and batch sizes. In particular, the PDD method exhibits a balanced approach in terms of both accuracy and the number of drift detections. For the Logistic Regression model, PDD achieves an accuracy of 0.6740 at 10 batches, which increases to 0.7349 at 20 batches and further to 0.7402 at 30 batches. Similarly, for the Decision Tree model, the accuracy values are 0.7253, 0.7392, and 0.7429 at 10, 20, and 30 batches, respectively. The Random Forest model shows accuracies of 0.7093, 0.7311, and 0.7449 across the three batch sizes.

% Elec2
\begin{table}[h]
    \centering
    \scriptsize
    \caption{The results of experiments on the \texttt{Elec2} dataset}
    \label{tab:exp_Elec2}
    \resizebox{\linewidth}{!}{
    \begin{tabular}{llcccccc}\toprule
         & & \multicolumn{2}{c}{Batch 10} & \multicolumn{2}{c}{Batch 20} & \multicolumn{2}{c}{Batch 30} \\\cmidrule(lr){3-4}\cmidrule(lr){5-6}\cmidrule(lr){7-8}
         \textbf{Method} & \textbf{Model} & \textbf{Acc.} & \textbf{\#drifts} & \textbf{Acc.} & \textbf{\#drifts} & \textbf{Acc.} & \textbf{\#drifts} \\\midrule
    % HDDM_A
    HDDM\_A & LR & 0.6983 & 10 & 0.6836 & 20 & 0.7099 & 30 \\
           & DT & 0.7255 & 10 & 0.7409 & 20 & 0.7450 & 30 \\
           & RF & 0.7029 & 10 & 0.7243 & 20 & 0.7374 & 30 \\\midrule
    % HDDM_W
    HDDM\_W & LR & 0.6983 & 10 & 0.6836 & 20 & 0.7099 & 30 \\
           & DT & 0.7255 & 10 & 0.7409 & 20 & 0.7450 & 30 \\
           & RF & 0.7029 & 10 & 0.7243 & 20 & 0.7374 & 30 \\\midrule
    % KSWIN
    KSWIN   & LR & 0.6983 & 10 & 0.6836 & 20 & 0.7099 & 30 \\
           & DT & 0.7255 & 10 & 0.7409 & 20 & 0.7450 & 30 \\
           & RF & 0.7029 & 10 & 0.7243 & 20 & 0.7374 & 30 \\\midrule
    % PageHinkley
    PH & LR & 0.6983 & 10 & 0.7342 & 13 & 0.7390 & 16 \\
                & DT & 0.7215 & 9  & 0.7417 & 15 & 0.7502 & 16 \\
                & RF & 0.6992 & 8  & 0.7103 & 11 & 0.7368 & 16 \\\midrule
    % DDM
    DDM     & LR & 0.6983 & 10 & 0.6836 & 20 & 0.7099 & 30 \\
            & DT & 0.7255 & 10 & 0.7409 & 20 & 0.7450 & 30 \\
            & RF & 0.7029 & 10 & 0.7250 & 16 & 0.7253 & 17 \\\midrule
    % EDDM
    EDDM    & LR & 0.6983 & 10 & 0.6836 & 20 & 0.7099 & 30 \\
            & DT & 0.7255 & 10 & 0.7409 & 20 & 0.7450 & 30 \\
            & RF & 0.7029 & 10 & 0.7243 & 20 & 0.7374 & 30 \\\midrule
    % PDD
    PDD     & LR & 0.6740 & 8  & 0.7349 & 4  & 0.7402 & 12 \\
            & DT & 0.7253 & 1  & 0.7392 & 1  & 0.7429 & 5  \\
            & RF & 0.7093 & 3  & 0.7311 & 0  & 0.7449 & 3  \\\bottomrule
    \end{tabular}}
\end{table}

When comparing drift detection counts, PDD maintains a moderate level of sensitivity. In the LR model, it detects 8 drifts at 10 batches, 4 at 20 batches, and 12 at 30 batches. For the DT model, the method is even more conservative, with just 1 drift detected at both 10 and 20 batches and an increase to 5 drifts at 30 batches. In the RF model, PDD identifies 3 drifts at 10 batches, no drift at 20 batches, and 3 drifts at 30 batches. This controlled drift detection contrasts with methods such as HDDM-A, HDDM-W, KSWIN, DDM, and EDDM, which consistently flag a higher number of drifts. In the LR model, these methods uniformly detect 10 drifts across all batch sizes, while the PH method shows an increasing trend with up to 16 drifts in the DT and RF models at 30 batches.

Accuracy remains fairly robust across all methods; the LR model maintains performance above 0.69, and the DT and RF models consistently achieve accuracies above 0.70. Such consistency in predictive performance is further underscored by the observation that the base test and train accuracies are high across all models. An additional key insight comes from the variable importance analysis: for the \texttt{Elec2} dataset, the attribute \texttt{nswprice} is identified as the most important variable for LR, DT, and RF models across all batch sizes. This stability in variable importance indicates that \texttt{nswprice} is a critical factor driving model predictions, which in turn may contribute to more reliable drift detection outcomes when this feature remains consistently influential.

Overall, the PDD method for the \texttt{Elec2} dataset demonstrates a balanced performance by maintaining competitive accuracy while detecting a moderate and controlled number of drifts. Its more conservative nature in comparison to other methods can be advantageous in operational scenarios where minimizing false positives is essential. Furthermore, the consistent identification of \texttt{nswprice} as the key variable across models reinforces the importance of focused variable analysis in understanding model behavior and drift dynamics.

\begin{figure}[h]
    \centering
    \includegraphics[width = \linewidth]{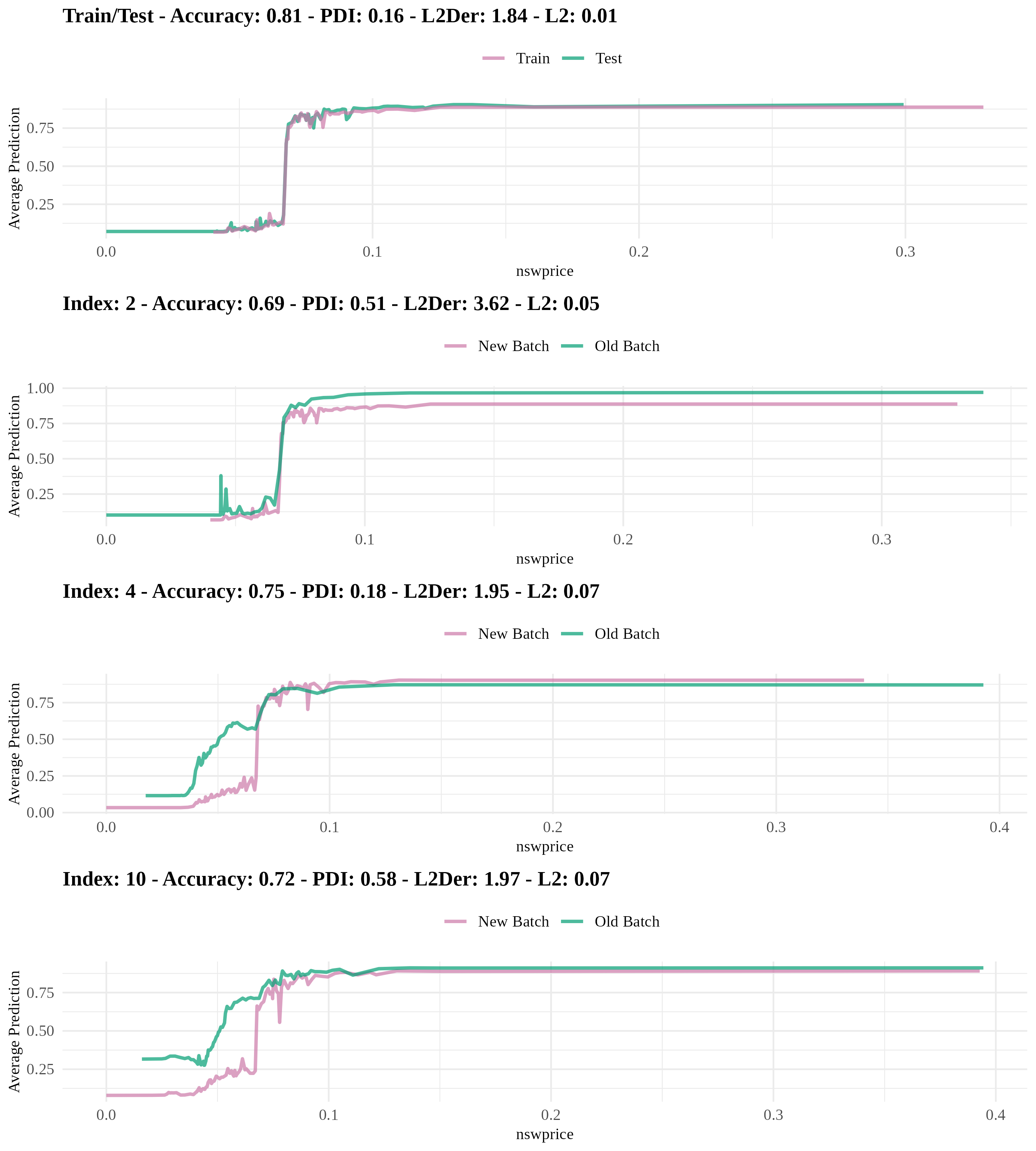}
    \caption{PDPs of the Random Forest model trained on \texttt{Elec2} dataset}
    \label{fig:rfelec10}
\end{figure}

For the RF model applied to the \texttt{Elec2} dataset, drift was detected in the $10$-th batch. In Figure~\ref{fig:rfelec10}, the \texttt{nswprice} variable, which is the most important variable, shows noticeable changes between $0$ and $0.1$. In this range, the contribution of \texttt{nswprice} to the model's predictions shifts, leading to drift detection by the PDD. These variations in \texttt{nswprice} highlight how its influence on the model differs in the $10$-th batch, potentially impacting the model’s performance.

%%%%%%%%%%%%%%%%%%%%%%%%%%%%%%%%%%%%%%%%%%%%%%%%%%%%%%%%%%%%%%%%%%%%%%%
\subsubsection{\texttt{Friedman}}

The results for the \texttt{Friedman} dataset in Table~\ref{tab:exp_Friedman}, which pertains to a regression task, exhibit distinct performance patterns across various drift detection methods, models, and batch sizes. Focusing on the PDD method, it demonstrated a balanced approach in terms of both RMSE values and the number of detected drifts. Specifically, for the LR model, PDD achieved RMSE values of $2.8332$, $2.6800$, and $2.7924$ across 10, 20, and 30 batches, respectively, while detecting 3, 2, and 1 drifts. In the DT model, PDD maintained stable RMSE values of $3.3460$, $3.1778$, and $3.2560$, with drift detections of $3$, $3$, and $2$ drifts across the same batch sizes. Similarly, for the RF model, the RMSE values reported by PDD are $2.0112$, $1.8544$, and $2.0467$, with drift detections of $8$, $18$, and $4$ drifts, respectively. This consistent performance indicates that PDD effectively maintains model accuracy while controlling the number of drift detections, thereby reducing the likelihood of unnecessary drift alarms.

% Friedman
\begin{table}[h]
    \centering
    \scriptsize
    \caption{The results of experiments on the \texttt{Friedman} dataset}
    \label{tab:exp_Friedman}
    \resizebox{\linewidth}{!}{
    \begin{tabular}{llcccccc}\toprule
 & & \multicolumn{2}{c}{Batch 10} & \multicolumn{2}{c}{Batch 20} & \multicolumn{2}{c}{Batch 30} \\\cmidrule(lr){3-4}\cmidrule(lr){5-6}\cmidrule(lr){7-8}
\textbf{Method} & \textbf{Model} & \textbf{RMSE} & \textbf{\#drifts} & \textbf{RMSE} & \textbf{\#drifts} & \textbf{RMSE} & \textbf{\#drifts} \\\midrule
\multirow{3}{*}{KSWIN} 
  & LR   & 2.8029 & 10   & 2.8007 & 19   & 2.7906 & 30   \\
  & DT   & 3.2977 & 10   & 3.2765 & 19   & 3.2921 & 23   \\
  & RF   & 1.9068 & 10   & 1.9282 & 20   & 1.9371 & 25   \\\midrule
\multirow{3}{*}{PH}
  & LR   & 2.8023 & 5    & 2.7974 & 10   & 2.7812 & 9    \\
  & DT   & 3.3254 & 9    & 3.2813 & 16   & 3.2752 & 16   \\
  & RF   & 1.9747 & 3    & 2.0654 & 5    & 2.0144 & 7    \\\midrule
\multirow{3}{*}{PDD}
  & LR   & 2.8332 & 3    & 2.6800 & 2    & 2.7924 & 1    \\
  & DT   & 3.3460 & 3    & 3.1778 & 3    & 3.2560 & 2    \\
  & RF   & 2.0112 & 8    & 1.8544 & 18   & 2.0467 & 4    \\\bottomrule
\end{tabular}}
\end{table}

In comparison, the KSWIN method consistently detected a higher number of drifts across all models and batch sizes. For instance, in the LR model, KSWIN identified $10$, $19$, and $30$ drifts with RMSE decreasing slightly from $2.8029$ to $2.7906$ as the number of batches increased. Similarly, in the DT and RF models, KSWIN detected $10$, $19$, and $23$ drifts and $10$, $20$, and $25$ drifts, respectively, with RMSE remaining relatively stable or showing minor improvements. This high sensitivity allows KSWIN to capture more drift events, which can be advantageous for detecting subtle changes in the data stream but may also lead to a higher risk of false positives.

The PH method exhibited moderate sensitivity, detecting fewer drifts compared to KSWIN but more than PDD. In the LR model, PH detected $5$, $10$, and $9$ drifts with RMSE improving from $2.8023$ to $2.7812$. The DT model showed consistent drift detections of $9$, $16$, and $16$ drifts, while the RF model detected $3$, $5$, and $7$ drifts across the batch sizes, all while maintaining stable RMSE values. PH strikes a balance between sensitivity and stability, making it suitable for scenarios where controlled drift detection is preferred without the excessive alerts that KSWIN might produce.

Across all methods and models, the most important variable consistently identified was \texttt{variable\_4}. This uniformity suggests that \texttt{variable\_4} plays a pivotal role in influencing model performance, regardless of batch size or the drift detection method employed. Such consistency underscores the importance of \texttt{variable\_4} in the predictive performance of the models and highlights the need for focused variable analysis in both drift detection and model evaluation.

In summary, the PDD method demonstrated balanced effectiveness for the Friedman regression task by maintaining reasonable RMSE values and controlling the number of detected drifts across different batch sizes. While PDD offers a more conservative approach to drift detection, reducing the likelihood of false positives, it may be less sensitive compared to methods like KSWIN, which, although more sensitive, carry a higher risk of false alarms. The PH method provides an intermediate level of sensitivity, offering a stable drift detection mechanism for controlled environments. Additionally, the consistent identification of \texttt{variable\_4} as the most important variable across all models and batch sizes emphasizes its critical role in model performance, suggesting that adaptive variable selection strategies could further enhance drift detection and overall model reliability in dynamic data environments.
%%%%%%%%%%%%%%%%%%%%%%%%%%%%%%%%%%%%%%%%%%%%%%%%%%%%%%%%%%%%%%%%%%%%%%%%%%%%%

\section{Conclusions}

In this paper, we conducted six experiments across multiple real-life and synthetic datasets to compare the effectiveness of our proposed method, PDD, and the most commonly used concept drift detection methods, including HDDM-A, HDDM-W, KSWIN, PH, DDM, and EDDM. We focused on analyzing model accuracy, the number of detected drifts, and understanding the variable contributions during drift occurrences.

PDD examines three metrics between two PDP curves: PDI, L2 (Euclidean distance), and L2Der (Euclidean distance between derivatives). By default, these three metrics are calculated from the PDPs derived from the train and test sets and are assigned as threshold values. For new incoming batches, these metrics from the test data are compared against the thresholds. In this study, the most important variable was selected based on variable importance using train set. Statistical tests also require adjusting their parameters, which can make them more or less sensitive depending on the type of data and drift. The same consideration applies to PDD. In addition to the default condition where all three metrics must exceed the thresholds, different conditions were also considered in the experiments, such as exceeding the threshold in at least two out of three metrics or exceeding the thresholds by a certain margin. In this study, statistical tests and PDD were compared in terms of the number of detected drifts and model performance across different batch sizes.

The main purpose of this paper is not only to detect concept drift but also to better understand the relationship between the model and variables when drift occurs. While traditional statistical methods are generally used to detect concept drift based on whether the response variable can be correctly predicted, PDD can be used for both detecting concept drift and gaining a deeper understanding of the changes. Additionally, since PDP does not rely on whether the response variable is correctly predicted, it was able to detect concept drift even at high accuracy levels. This allows for detecting changes when the relationships in the data change, but no change is observed in the model performance metrics, such as accuracy, F1, etc., thereby enabling the model to be updated accordingly.

The PDD method, a primary focus of our analysis, demonstrated competitive performance in terms of both accuracy and drift detection consistency. It effectively balanced the detection of drifts while maintaining high accuracy across multiple datasets, particularly when compared to more sensitive methods like KSWIN and EDDM, which frequently detected a higher number of drifts, including potential false positives. PDD's conservative approach to drift detection proved beneficial in controlling unnecessary drift alarms, which is crucial for maintaining stability in real-world applications.

In the \texttt{SEA} and \texttt{Elec2} datasets, PDD showed stable performance by controlling the number of drift detections while achieving high accuracy rates. The experiments on the \texttt{Hyperplane} dataset highlighted PDD's sensitivity to batch size, where performance decreased beyond the 20-batch level, likely due to overfitting effects. This limitation was also observed in the \texttt{NOAA} and \texttt{Ozone} datasets, where overfitting in the RF model impacted the PDD's drift detection capability. However, in the \texttt{Friedman} dataset, PDD maintained a good balance between detecting significant drifts and minimizing false positives, providing consistent metric values across different models and batch sizes.

Comparatively, methods like KSWIN and EDDM exhibited higher sensitivity to concept drift, detecting a greater number of drifts across all experiments. While this high sensitivity is useful for capturing real changes in the data stream, it also leads to an increased risk of false positives, which may be undesirable in operational settings. On the other hand, PH and DDM provided moderate performance, striking a balance between drift detection and stability without generating excessive alarms.

An important observation across all datasets was the consistency in identifying key variables that contributed to model predictions and drift detection. For instance, attributes like \texttt{V2} in \texttt{SEA}, \texttt{nswprice} in \texttt{Elec2}, and \texttt{variable\_4} in the \texttt{Friedman} dataset were repeatedly identified as the most influential variables. This consistency emphasizes the importance of monitoring these key attributes, as their behavior significantly affects model performance during concept drift.
%%%%%%%%%%%%%%%%%%%%%%%%%%%%%%%%%%%%%%%%%%%%%%%%%%%%%%%%%%%%%%%%%%%%%%%%%%%%%

\section{Discussion}

Except for the detection ability of our proposed method, it can provide explanations to understand the reason for predictive churn. After drift detection, re-training ML models can unpredictably change certain model predictions \citep{daniels_2024}. This may lead to unreliable conclusions about the model predictions. Fortunately, PDD can provide more detailed information about the relationship between the model and the data compared to other statistical methods in such situations. In this way, model users can understand the reason behind the drift. Similarly, users may request the deletion of their data regarding data privacy, and removing these relationships from any model trained on this data is a challenging problem. \cite{bourtoule_2021} developed a method to address this issue by partitioning and ordering the data. PDD can contribute to such problems in the literature.

The PDD method has some disadvantages, such as computational cost and infeasible use for multi-class classification. For example, the calculation of derivatives can become challenging for categorical variables with a small number of levels, which may prevent PDD from calculating its metrics. By default, PDD relies on the most important variable in the model, but when the number of explanatory variables increases or when their contribution to the response variable grows, detecting concept drift based solely on the most important variable may overlook drifts caused by other variables. In such cases, it is necessary to detect concept drift based on other explanatory variables as well, which increases the computational cost of PDD. In situations with many explanatory variables, the most important variable may need to be updated for each batch. PDD can also be more affected by small changes in sparse vectors because the three metrics calculated for PDD are averaged across all selected observation points. For example, the PDI metric measures whether the directions of derivatives are the same. For vectors with different derivative directions in 5 out of 100 observations, the PDI is calculated as 0.05, whereas for vectors with different directions in 1 out of 5 observations, the PDI increases to 0.20. 

On the other hand, the PDD method can work on binary classification or regression tasks. When the response variable has more than two classes, PDD can be adapted for a specific level of the response variable, but this adaptation and tracking can be costly. 

In conclusion, PDD offers advantages and disadvantages compared to traditional concept drift detection methods. The experiments demonstrated that PDD generally detected fewer but more consistent drifts while maintaining similar accuracy rates to other tests. Additionally, PDD allows for a more detailed examination of the relationship between explanatory variables and the response variable in the model, providing valuable insights into how these relationships change during concept drift.
%%%%%%%%%%%%%%%%%%%%%%%%%%%%%%%%%%%%%%%%%%%%%%%%%%%%%%%%%%%%%%%%%%%%%%%%%%%%%

\section*{Further research}

Reducing the computational cost of the PDD method, the \textit{compress then explain} \citep{hubert_2024} technique can be integrated. This technique compresses the data distribution when the number of observations is large and subsequently applies explainable methods more effectively. However, in cases where model training costs are high, it can be used in its current form. Additionally, when the cost of training the model is lower than the cost of generating PDPs, it can be used to monitor how the relationship between the response variable and explanatory variable changes when concept drift is detected in the model. Moreover, the PDD can be used in predictive churn and data privacy applications to make the changes explainable after re-training an ML model. 
%%%%%%%%%%%%%%%%%%%%%%%%%%%%%%%%%%%%%%%%%%%%%%%%%%%%%%%%%%%%%%%%%%%%%%%%%%%%%

\section*{CRediT authorship contribution statement}
\textbf{Ugur Dar:} Writing – original draft, Visualization, Software, Methodology, Investigation, Funding acquisition; \textbf{Mustafa Cavus:} Conceptualization, Writing – review \& editing, Supervision.

\section*{Data and code availability}
The materials for reproducing the experiments performed, and the dataset are accessible in the following repository: \url{https://github.com/ugurdar/datadriftR_DMKD}.

\section*{Declaration of competing interest}
The authors declare that they have no known competing financial interests or personal relationships that could have influenced the work reported in this paper.

\section*{Acknowledgements}
The work on this paper is financially supported by the Scientific and Technological Research Council of Turkiye under the 2210C National MSc Scholarship Program in the Priority Fields in Science and Technology, grant no. 1649B022303919 and the Eskisehir Technical University Scientific Research Projects Commission under grant no. 22LÖT175.
%%%%%%%%%%%%%%%%%%%%%%%%%%%%%%%%%%%%%%%%%%%%%%%%%%%%%%%%%%%%%%%%%%%%%%%%%%%%%

\appendix
\section{The description of the batches}

\label{sec:app}

\begin{table}[H]
    \centering
    \small
    \caption{The number of observations in the batches}
    \label{tab:batch_number}
    \resizebox{\linewidth}{!}{
    \begin{tabular}{lrrrrrr}\toprule
                & \multicolumn{6}{c}{\textbf{\#batch}}\\\cmidrule(lr){2-7}
                & \multicolumn{2}{c}{\textbf{10}} & \multicolumn{2}{c}{\textbf{20}} & \multicolumn{2}{c}{\textbf{30}} \\\cmidrule(lr){2-7}
        \textbf{Dataset} & \textbf{train} & \textbf{test} & \textbf{train} & \textbf{test} & \textbf{train} & \textbf{test} \\\midrule
        \texttt{SEA} & 13332 & 3334 & 8000 & 2000 & 5713 & 1429 \\
        \texttt{Hyperplane} & 5332 & 1334 & 3200 & 800 & 2285 & 572 \\
        \texttt{NOAA} & 4842 & 1211 & 2904 & 727 & 2075 & 519 \\
        \texttt{Ozone} & 675 & 169 & 404 & 102 & 289 & 73 \\
        \texttt{Elec2} & 12083 & 3021 & 7249 & 1813 & 5178 & 1295 \\
        \texttt{Friedman} & 5332 & 1334 & 3200 & 800 & 2285 & 572 \\
        \bottomrule
    \end{tabular}}
\end{table}

\begin{table}[H]
    \centering
    \small
    \caption{The most important variables in the batches}
    \label{tab:vip_batch}
    \resizebox{\linewidth}{!}{
    \begin{tabular}{llccc}\toprule
                            &       & \multicolumn{3}{c}{\textbf{\#batch}}\\\cmidrule(lr){3-5}
        \textbf{Dataset}    & \textbf{Model} & \textbf{10}                    & \textbf{20}                    & \textbf{30} \\\midrule
        \texttt{SEA} & RF & \texttt{V2} & \texttt{V2} & \texttt{V2} \\
                            & LR & \texttt{V2} & \texttt{V2} & \texttt{V2} \\
                            & DT & \texttt{V1} & \texttt{V1} & \texttt{V1} \\
        \midrule
        \texttt{Hyperplane} & RF & \texttt{feature\_1} & \texttt{feature\_1} & \texttt{feature\_1} \\
                            & LR & \texttt{feature\_1} & \texttt{feature\_1} & \texttt{feature\_1} \\
                            & DT & \texttt{feature\_1} & \texttt{feature\_1} & \texttt{feature\_2} \\
        \midrule
        \texttt{NOAA} & RF & \texttt{attribute2} & \texttt{attribute4} & \texttt{attribute2} \\
                            & LR & \texttt{attribute8} & \texttt{attribute4} & \texttt{attribute2} \\
                            & DT & \texttt{attribute2} & \texttt{attribute2} & \texttt{attribute2} \\
        \midrule
        \texttt{Ozone} & RF & \texttt{V56} & \texttt{V61} & \texttt{V61} \\
                            & LR & \texttt{V25} & \texttt{V19} & \texttt{V19} \\
                            & DT & \texttt{V31} & \texttt{V36} & \texttt{V36} \\
        \midrule
        \texttt{Elec2} & RF & \texttt{nswprice} & \texttt{nswprice} & \texttt{nswprice} \\
                            & LR & \texttt{nswprice} & \texttt{nswprice} & \texttt{nswprice} \\
                            & DT & \texttt{nswprice} & \texttt{nswprice} & \texttt{nswprice} \\
        \midrule
        \texttt{Friedman} & RF & \texttt{feature\_4} & \texttt{feature\_4} & \texttt{feature\_4} \\
                            & linear & \texttt{feature\_4} & \texttt{feature\_4} & \texttt{feature\_4} \\
                            & DT & \texttt{feature\_4} & \texttt{feature\_4} & \texttt{feature\_4} \\
        \midrule
    \end{tabular}}
\end{table}

\begin{table}[H]
    \centering
    \scriptsize
    \caption{The accuracies of the models in the batches}
    \label{tab:acc_batch}
    \resizebox{\linewidth}{!}{
    \begin{tabular}{llcccccc}\toprule
    & & \multicolumn{6}{c}{\textbf{\#batch}}\\
    \cmidrule(lr){3-8}
    & & \multicolumn{2}{c}{\textbf{10}} & \multicolumn{2}{c}{\textbf{20}} & \multicolumn{2}{c}{\textbf{30}} \\
    \cmidrule(lr){3-8}
    \textbf{Dataset} & \textbf{Model} & \textbf{test} & \textbf{train} & \textbf{test} & \textbf{train} & \textbf{test} & \textbf{train} \\
    \midrule
    \texttt{SEA} & RF & 0.8537 & 0.9991 & 0.8896 & 0.9989 & 0.9007 & 0.9991 \\
                          & LR & 0.8579 & 0.8966 & 0.8946 & 0.9032 & 0.9007 & 0.9057 \\
                          & DT & 0.8576 & 0.8801 & 0.8696 & 0.8850 & 0.8797 & 0.8850 \\
    \midrule
    \texttt{Hyperplane} & RF & 0.6524 & 0.9992 & 0.6692 & 0.9984 & 0.4799 & 0.9978 \\
                          & LR & 0.6165 & 0.7320 & 0.6717 & 0.7122 & 0.5113 & 0.7580 \\
                          & DT & 0.7079 & 0.7406 & 0.7266 & 0.7094 & 0.4276 & 0.7575 \\
    \midrule
    \texttt{NOAA} & RF & 0.7723 & 0.9996 & 0.8338 & 0.9997 & 0.8154 & 0.9995 \\
                          & LR & 0.7748 & 0.7999 & 0.8118 & 0.8041 & 0.8173 & 0.8014 \\
                          & DT & 0.7500 & 0.7881 & 0.7734 & 0.7996 & 0.7808 & 0.7990 \\
    \midrule
    \texttt{Ozone} & RF & 0.9941 & 1.0000 & 0.9806 & 1.0000 & 1.0000 & 1.0000 \\
                          & LR & 0.9647 & 0.9526 & 0.8155 & 1.0000 & 0.9054 & 1.0000 \\
                          & DT & 0.9882 & 0.9511 & 0.8544 & 0.9554 & 0.9730 & 0.9308 \\
    \midrule
    \texttt{Elec2} & RF & 0.8107 & 0.8424 & 0.8142 & 0.8367 & 0.8040 & 0.8418 \\
                          & LR & 0.8154 & 0.8200 & 0.8043 & 0.8081 & 0.8156 & 0.8188 \\
                          & DT & 0.7955 & 0.8264 & 0.8649 & 0.8238 & 0.8025 & 0.8316 \\
    \midrule
    \texttt{Friedman} & RF & 1.6639 & 0.7431 & 1.8074 & 0.7709 & 1.7661 & 0.7949 \\
                          & linear & 2.6432 & 2.5926 & 2.5591 & 2.5579 & 2.5958 & 2.5463 \\
                          & DT & 3.0254 & 2.9843 & 3.1574 & 2.9204 & 3.0400 & 2.9744 \\
    \midrule
    \end{tabular}}
\end{table}

%%       \citet{<label>} ==> Lamport (1994)
%%       \citep{<label>} ==> (Lamport, 1994)
%%
%Example citation, See \citet{lamport94}.




%\bibliographystyle{elsarticle-harv} 
%\bibliography{cas-refs}

%\bibitem[Lamport(1994)]{lamport94}
%  \textit{\LaTeX: a document preparation system},
%  Addison Wesley, Massachusetts,

\end{document}